\documentclass[12pt,english]{article}
\usepackage[latin9]{inputenc}
\usepackage{booktabs}
\usepackage{mathtools}
\usepackage{bm}
\usepackage{multirow}
\usepackage[linesnumbered,ruled,vlined]{algorithm2e}
\usepackage{amsmath}
\usepackage{amsthm}
\usepackage{amssymb}
\usepackage{graphicx}
\usepackage{tablefootnote}
\usepackage{setspace}
\usepackage[authoryear,round,sort]{natbib}
\usepackage[unicode=true,
 bookmarks=false,
 breaklinks=true,pdfborder={0 0 1},backref=false,colorlinks=false]
 {hyperref}
\hypersetup{
 pdfborderstyle=}

\makeatletter

\DeclareTextSymbolDefault{\textquotedbl}{T1}
\providecommand{\tabularnewline}{\\}

\theoremstyle{plain}
\newtheorem{thm}{\protect\theoremname}
\theoremstyle{plain}
\newtheorem{prop}[thm]{\protect\propositionname}
\theoremstyle{definition}
\newtheorem*{example*}{\protect\examplename}

\@ifundefined{date}{}{\date{}}
\usepackage{multirow}
\usepackage{tablefootnote}
\usepackage{thmtools}
\usepackage{psfrag}\usepackage{epsf}\usepackage{enumerate}

\usepackage{adjustbox}
\usepackage{placeins}
\usepackage[pagewise]{lineno}\linenumbers
\modulolinenumbers[5]
\nolinenumbers

\newcommand{\st}[1]{\ignorespaces}
\providecommand{\propositionname}{Proposition}
\providecommand{\theoremname}{Theorem}

\usepackage{xcolor}

\g@addto@macro\@floatboxreset\centering

\newcommand{\hlnote}[1]{{\color{green}[ #1 - HL ]}}

\newcommand{\xslnote}[1]{{\bf\color{blue}[ #1 -- SHERRY ]}}

\newcommand{\ycnote}[1]{{\color{magenta}[ #1 - YC ]}}
\newcommand{\ycrev}[1]{{\color{black}#1}}

\newcommand{\ylrevnew}[1]{{\color{black}#1}}
\newcommand{\hlrev}[1]{{\color{black}#1}}

\usepackage{amsfonts}
\usepackage{epstopdf}
\usepackage{algorithmic}
\usepackage{enumitem}

\makeatother

\providecommand{\examplename}{Example}
\providecommand{\propositionname}{Proposition}
\providecommand{\theoremname}{Theorem}

\begin{document}
\title{\textbf{Hybrid Parameter Search and Dynamic Model Selection for Mixed-Variable
Bayesian Optimization}}
\author{Hengrui Luo (hrluo@lbl.gov),\thanks{Lawrence Berkeley National Laboratory, Berkeley, CA, 94701.}\\
 Younghyun Cho (younghyun@berkeley.edu)\thanks{University of California, Berkeley, Berkeley, CA, 94720.},\\
 James W. Demmel (demmel@berkeley.edu) $^{\dagger}$,\\
 Xiaoye S. Li (xsli@lbl.gov)$^{*}$,\\
 Yang Liu (liuyangzhuan@lbl.gov)$^{*}$\\
 }
\maketitle
\begin{abstract}
This paper presents a new type of hybrid model for Bayesian optimization (BO) adept at managing mixed variables, encompassing both quantitative (continuous and integer) and qualitative (categorical) types. Our proposed new hybrid models (named hybridM) merge the Monte Carlo Tree Search structure (MCTS) for categorical variables with Gaussian Processes (GP) for continuous ones. \ylrevnew{hybridM leverages} the \st{original (frequentist)} upper confidence bound tree search (UCTS) for MCTS 
\st{and the Bayesian Dirichlet search} strategy, showcasing the tree architecture's integration into Bayesian optimization. \st{Central to our innovation in surrogate modeling phase is online kernel selection for mixed-variable BO.} Our innovations, including dynamic \ylrevnew{online} kernel selection in the surrogate modeling phase and a unique UCTS search strategy\st{and Bayesian update strategies (hybridD)}, position our hybrid models as an advancement in mixed-variable surrogate models. Numerical experiments underscore the superiority of hybrid models, highlighting their potential in Bayesian optimization.
\end{abstract}
\textit{Keywords:} Gaussian processes, Monte Carlo tree search, categorical variables, online kernel selection.

\section{\label{sec:Introduction}Introduction}

Our motivating problem is to optimize a ``black-box''
function with ``mixed'' variables, without
an analytic expression. ``Mixed'' signifies that
the function's input variables comprise both continuous (quantitative) and categorical (qualitative) 
variables, common in machine learning and scientific computing tasks
such as the performance tuning of mathematical
libraries and application codes at runtime and compile-time \citep{balaprakash2018autotuning}.

Bayesian optimization (BO) with Gaussian process (GP) surrogate models
is a common method for optimizing noisy and expensive black-box functions,
designed primarily for continuous-variable functions \citep{shahriari_taking_2016,sid2020gptune}.
\hlrev{In BO we draw (possibly noisy) sequential samples from black-box functions at any query points. Then we use a GP surrogate model fitted with sequential samples $(\bm{x},y)$ chosen by optimizing the criteria based on surrogate.
Since optimizing criteria (e.g.,} expected improvement by \citet{jones1998efficient}) \hlrev{is defined on a  continuous-variable domain, e}xtending BO to mixed-variable functions presents challenges due to variable type differences (Table \ref{tab:Comparison-between-different}).

Continuous variables have uncountably many values with magnitudes
and intrinsic ordering, allowing natural gradient definition. In contrast, categorical variables, having finitely many
values without intrinsic ordering or magnitude, require encoding in
the GP context, potentially inducing discontinuity and degrading GP
performance \citep{luo_nonsmooth_2021}.
In addition to continuous and categorical variables, \textit{integer
variables} (a.k.a. ordinal variables with intrinsic
ordering) have discrete values, natural ordering and discrete gradients.
The empirical rule of thumb for handling an integer variable \citep{karlsson2020continuous}
is to treat it as a categorical variable if the number of integer
values (i.e., number of categorical values) is small, or as a continuous
variable with embedding (a.k.a. rounding-off encoding) otherwise.
We follow this empirical rule for integer variables\hlrev{, and treat mixed-variable problems which contains all these kinds of variables.} 
\begin{table}
\begin{centering}
\begin{tabular}{|c|c|c|c|}
\hline 
Type  & \textbf{Continuous }  & \textbf{Integer }  & \textbf{Categorical }\tabularnewline
\hline 
\hline 
Representation  & $\mathbb{R}$  & $\mathbb{Z}$  & finite set \tabularnewline
\hline 
Magnitude  & Yes  & Yes  & No\tabularnewline
\hline 
Order  & Yes  & Yes  & No\tabularnewline
\hline 
Encoding  & No  & Yes  & Yes\tabularnewline
\hline 
Gradient  & continuous gradient  & discrete gradient  & No\tabularnewline
\hline 
\end{tabular}
\par\end{centering}
\caption{\label{tab:Comparison-between-different}Comparison among different
types of variables. The discussion of different types of encodings
can be found in \citet{cerda2018similarity}.}
\end{table}

Existing strategies for addressing challenges in mixed-variable Bayesian
Optimization (BO) can be categorized into four non-exclusive classes.

First, encoding transforms categorical variables into numerical format, enabling algorithms to process and analyze non-numeric data effectively. Encoding methods such as one-hot encoding (representing categories
with binary numbers \citep{snoek_practical_2012}) and graph-based
encoding (using graph vertices to represent categories \citep{karlsson2020continuous}),
are widely used for mixed variables. However, their theoretical justification
and practical performance are highly problem-specific \citep{cerda2018similarity,garrido-merchan_dealing_2020},
motivating more advanced mixed-variable techniques.

Second, new acquisition functions have been designed to address specific
problems in discrete or categorical space \citep{willemsen_bayesian_2021,deshwal2021bayesian,oh2021mixed}.
These \ylrevnew{advanced} designs preserve the BO framework and the Gaussian
Process (GP) surrogate. But these new acquisition functions are often
difficult to optimize and require special implementations \hlrev{that may introduces further optimization challenges in the inner loop}. 

Third, \ylrevnew{modification of the surrogate has led to new GP models} such as additive GP models \citep{deng_additive_2017,xiao2021ezgp}, latent variable GP \citep{zhang2020bayesian,zhang2020latent}, 
models based on non-GP types of basis functions \citep{bliek2021black},
graphical models \citep{olson_tpot_2019,headtim_scikitoptimize_2020,biau2021optimization},
and models with novel kernels capturing the categorical variables
using a density estimator (TPE, \citep{bergstra2011algorithms}) or
random forests (SMAC, \citep{hutter_sequential_2011}). \hlrev{These novel
surrogate models are kind of generic modeling
strategies that can be used for mixed-variable input space}, \ycrev{however, they} usually face challenges in higher-dimensional
problems \hlrev{with a lot of categories}.

Fourth, bandit-based search methods utilize either multiarmed bandits
(MAB) (EXP3BO \citep{gopakumar_algorithmic_2018}, Bandit-BO \citep{nguyen_bayesian_2019},
CoCaBO \citep{ru_bayesian_2020}), or Monte Carlo tree search (MCTS)
(MOSAIC, \citep{rakotoarison_automated_2019}) to select the optimal
categorical variables. The model then fits the remaining continuous
variables using a standard GP surrogate that includes all of these
variables. However, the regular bandit (EXP3BO, BanditBO, CoCaBO)
needs to construct all possible combinations of categories and cannot
handle large number of categories. On the other hand, the tree-like bandit (MOSAIC) 
is more scalable for large number of combinations. However, MOSAIC requires independent GPs and a large budget per GP to behave reasonably.
\hlrev{For problems with many categories, this approach can also be inefficient and overlooks interactions between continuous and categorical variables.}

\begin{table}
\begin{centering}
\begin{tabular}{|c|c|c|}
\hline 
 & Categorical  & Continuous\tabularnewline
\hline 
\hline 
\multicolumn{3}{|c|}{\textbf{Hybrid models}}\tabularnewline
\hline 
hybridM (proposed)  & MCTS   & dependent GPs\tabularnewline
\hline 
MOSAIC \citep{rakotoarison_automated_2019}  & MCTS  & independent GPs\tabularnewline
\hline 
GP (skopt) \citep{headtim_scikitoptimize_2020}  & sampling/encoding  & independent GPs\tabularnewline
\hline 
CoCaBO \citep{ru_bayesian_2020}  & 1-layer bandit  & dependent GPs\tabularnewline
\hline 
EXP3BO\tablefootnote{This includes the implementation for roundrobin MAB and random MAB.}
\citep{gopakumar_algorithmic_2018}  & 1-layer bandit  & independent GPs\tabularnewline
\hline 
\multicolumn{3}{|c|}{\textbf{Non-hybrid models}}\tabularnewline
\hline 
SMAC \citep{hutter_sequential_2011}  & {*}  & {*} \tabularnewline
\hline 
TPE \citep{bergstra2011algorithms}  & tree estimator  & GPs\tabularnewline
\hline 
forest (skopt) \citep{headtim_scikitoptimize_2020}  & tree estimator  & tree estimator\tabularnewline
\hline 
\end{tabular}
\par\end{centering}
\caption{\label{tab:Model-unification-via}Comparison among different mixed-variable
surrogate models. Dependent GPs refer to GPs with mixed variables,
while independent GPs refer to GPs with only continuous variables.\\
{*}For SMAC, we use SMAC3 (1.2.0) and refer to their paper \citep{hutter_sequential_2011} for the details of their complex model.}
\end{table}

\hlrev{Our proposed \textit{hybrid}} \ylrevnew{\textit{model} (henceforth dubbed hybridM)} uses tree structure for categorical space
and GP for continuous space. 
hybridM presents a ``tree-and-GP'' 
architecture for search phase and utilizes
a \ylrevnew{novel \textit{dependent}} Gaussian Process (GP) surrogate with a selection of mixed-variable
kernels to capture correlations between continuous and \hlrev{categorical
variables (See Figure \ref{fig:Schema-hybrid}) for modeling phase. In addition, our unique approach integrates a dynamic kernel selection strategy, balancing optimism} \citep{luozhu_opt,ye1998measuring} \hlrev{and likelihood. This method enhances the efficiency of model selection and optimization, effectively addressing the limitations of the MOSAIC large tuning sample budget requirement.}

\hlrev{
This approach is our first attempt to the open problem of conducting model selection in an online context}\st{, a challenge that motivates our model construction}. \st{Notably, our use of 'optimism' (highest acquisition) and a combined ranking system for kernel selection based on both optimism and likelihood is not limited to mixed discrete-continuous cases. This versatility presents substantial potential for future research.} \hlrev{The overall workflow of the proposed method is as follows}.
We first employ tree structures for heuristic search of the next
sample's categorical part with the upper confidence bound tree search (UCTS) 
strategy \st{ or the novel Bayesian Dirichlet-Multinomial strategy }(see Sections \ref{subsec:Monte Carlo-Tree-Search}
 and Appendix \ref{sec:Bayesian Update Strategies}). 
We then dynamically select
the optimal kernel for the mixed-variable GP surrogate \ylrevnew{(see Section \ref{sec:kernel selection})}.
Finally, we use acquisition maximization to search for the next sample's continuous
variables using the best continuous GP kernel, conditioned on the
categorical variables.


Our contributions are summarized below. 
\begin{itemize}
\item We develop a novel hybrid model with a tree and \ylrevnew{dependent} GP structure \hlrev{integrated with MCTS} \st{which
integrates the UCTS and Bayesian search (hybridD) strategy,} extending the state-of-the-art architectures in Table \ref{tab:Model-unification-via}.
\item 
We experiment with existing kernels for mixed-variable situations
and introduce a family of candidate kernels for the mixed-type variable
BO (Appendix \ref{subsec:Kernel-Design}).
\item We propose
a numerically stable and efficient dynamic kernel selection criterion
(Section \ref{subsec:Kernel-Selection}) to enhance tuning performance.
\item 
We demonstrate the superior tuning performance of our hybrid model through various synthetic benchmarks as well as real application codes, including tunings of a neural network and a parallel sparse solver STRUMPACK\citep{doecode_37094}.
\end{itemize}

Tuning scientific applications like  neural networks involves optimizing numerous different types of hyperparameter \citep{elsken2019neural}, which affects performance outcomes. 
For another example, optimizing  STRUMPACK \citep{doecode_37094}, a software package for solving large sparse linear systems, is rather expensive to tune due to the time cost per execution 
with certain configurations and its data-dependent nature.
Both applications present unique challenges and require efficient tuning, 
emphasizing the need for effective auto-tuning strategies. Comprehensive
benchmarks for mixed-variable methods are provided in Section \ref{sec:Experiments},
which showcasing the superiority of our hybrid models in diverse scenarios.

\section{\label{sec:Hybrid-Models}New Hybrid Model}

\begin{figure}[t]
\begin{centering}
\includegraphics[scale=0.4]{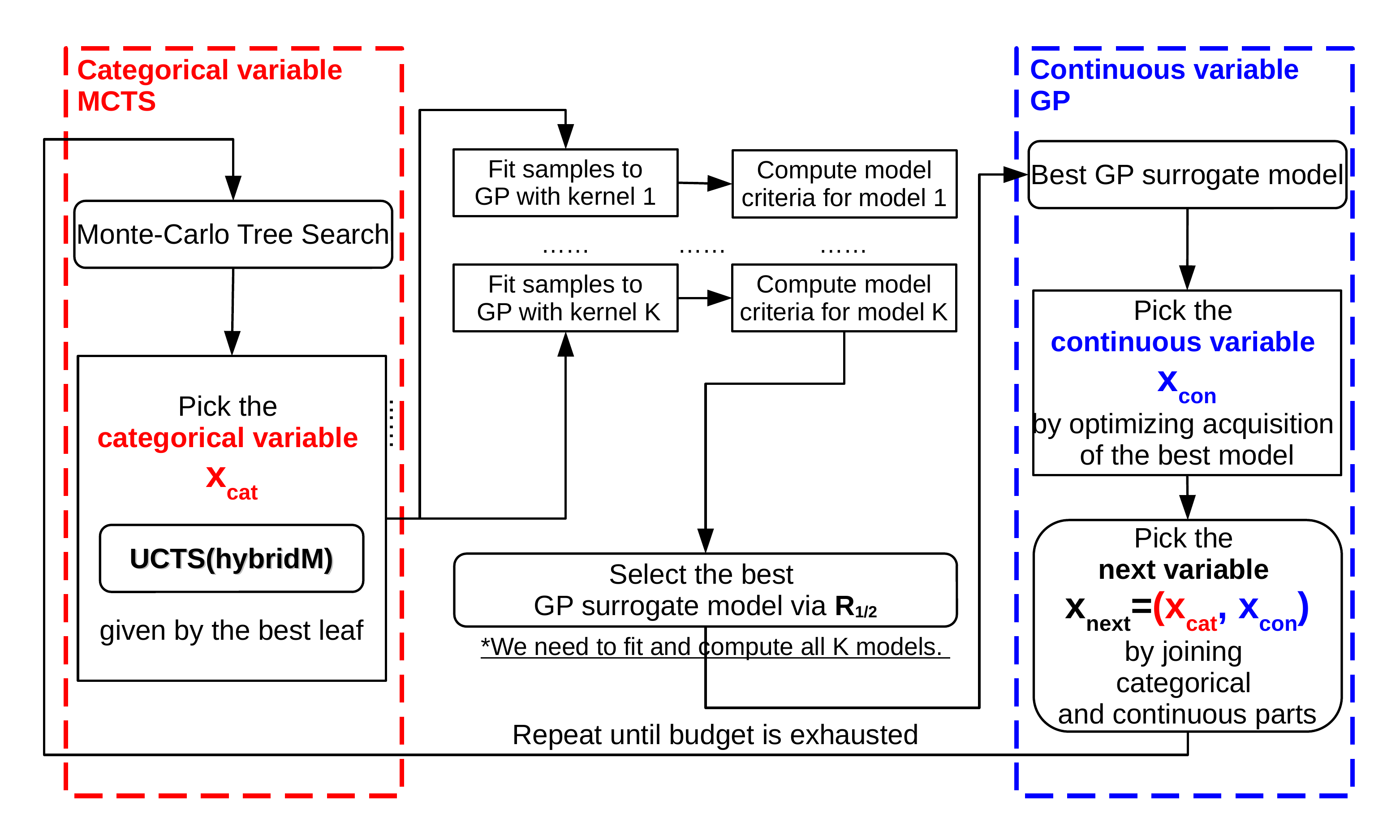} 
\par\end{centering}
\caption{\label{fig:Schema-hybrid}Algorithmic representation for the update
step (i.e., heuristic search for the next sequential sample) in the
proposed hybrid model, where we provide two different stretegies
UCTS (hybridM) \st{and Bayesian (hybridD)}. Note that the fitting of GPs with different kernels
can be parallelized rather than fitted sequentially. }
\end{figure}

This section describes our hybrid methodology, which is
illustrated by Figure \ref{fig:Schema-hybrid} and detailed later in Algorithm
\ref{alg:hybrid}. 
The section is organized as follows: In Section \ref{subsec:Monte Carlo-Tree-Search},
we first introduce the MCTS, which will be used to decide the categorical
part of the next sequential sample when used with UCTS (alternatively, we provide Bayesian strategies in Appendix \ref{sec:Bayesian Update Strategies}).
With GP surrogate and
families of covariance kernels (detailed in Appendix \ref{subsec:Kernel-Design}), which will be used to decide the continuous
part of the next sequential sample, we search the categorical part via a tree strategy and the continuous part via a GP strategy.  
In Section \ref{sec:kernel selection}, after deciding both the
categorical and continuous part of the variable,
we explain how to choose the mixed-variable kernel
for the GP, which models the overall variable (including both categorical and continuous part via selected mixed kernels) and gives us \ycrev{a} balance
between exploration and exploitation. 
In Section~\ref{sec:put_together}, we summarize our hybridM algorithm when all components are combined.

\subsection{\label{subsec:Monte Carlo-Tree-Search}UCTS Update Strategy}

UCTS is one of the MCTS search strategies that constructs a tree structure
as the Bayesian Optimization (BO) sampling proceeds. Each sequential
sample is composed of categorical and continuous parts, denoted as
$\bm{x}=(\bm{x}_{\text{cat}},\bm{x}_{\text{con}})$ \hlrev{ and the totality of responses as $\bm{y}$.}

In the MCTS methodology, the tree structure represents the categorical variables $n_{c}$ of $\bm{x}_{\text{cat}}$. Nodes at the same level represent
different choices of categorical values for the corresponding variable.
If a tree has $L\leq n_{c}$ levels, then each level $i$ has
$K_{i}$ nodes. The tree has a total of $C$ combinations of categorical
values, represented by $C$ leaves, where $C=\prod_{i}K_{i}$. When
$L=n_{c}$, each layer of the tree corresponds to a categorical
variable. Often, one can prune nodes and collapse layers of the tree
to reduce the categorical search space. Also, the tree can respect the sequential ordering of
the variables by using parent nodes for choice of an algorithm
and children nodes for algorithm-specific parameters.
It is worth mentioning that we choose MCTS instead of the
bandit-based algorithms (e.g., CoCaBO and EXP3BO in Table \ref{tab:Model-unification-via})
and sampling-based algorithm (e.g., skopt in Table \ref{tab:Model-unification-via})
because bandit- and sampling-based algorithms perform poorly for large $C$
(see Section 5.2 of \citet{ru_bayesian_2020}).

The workflow of MCTS \citep{kocsis_bandit_2006} is summarized as
follows: Consider a complete path $\bm{s}=(s_{1},s_{2},\cdots,s_{L})$
of length $L$, where $s_{1},s_{2},\cdots,s_{L}$ represent the nodes
(i.e., categorical values) at each level. This complete path corresponds
to a full categorical part of the variable $\bm{x}_{\text{cat}}$
and can be evaluated to generate a reward function $r(\bm{s})$. To
be precise, we can consider a continuous-variable GP conditioned on
the path $\bm{s}$, i.e., the corresponding categorical value. When maximizing a function $f$, we use the black-box function $r(\bm{s})=f(\bm{s},\bm{x}_{\text{con}}^{*})$
as the reward value to be maximized, following \citet{ru_bayesian_2020,rakotoarison_automated_2019}. We can use negative function values when minimizing the black-box function. The notation $\bm{x}_{\text{con}}^{*}$ is the new continuous
variable value proposed by the maximization of the acquisition of the surrogate
corresponding to the leaf node. In other words, a new function sample
is generated to evaluate $\bm{s}$, see Line 13 of Algorithm
\ref{alg:hybrid}.

Since the same path $\bm{s}$ can be visited multiple times during
the search, we use the mean of the historically generated $r(s_{1},s_{2},\cdots,s_{L})$
as the averaged reward function $\bar{r}(s_{1},s_{2},\cdots,s_{L})$
for the leaf nodes. For each non-leaf interior node $s_{i}$, $i<L$,
the reward function $\bar{r}(s_{1},s_{2},\cdots,s_{i})$ is simply
the average of $\bar{r}(s_{1},s_{2},\cdots,s_{i},s_{i+1})$ for all
$s_{i+1}$ visited so far. Finally, we use the maximization of the
following function, namely the upper confidence bound (UCB) in \citep{auer2002using},
as the \emph{upper confident bound tree search (UCTS) policy}, to
search for new categorical variable values: 
\begin{equation}
\bar{r}(s_{1},s_{2},\cdots,s_{i})+C_{UCB}\sqrt{\frac{\log n(s_{1},s_{2},\cdots,s_{i-1})}{n(s_{1},s_{2},\cdots,s_{i-1},s_{i})}},\label{eq:UCB policy}
\end{equation}
where the first and second terms serve respectively as the exploitation
and exploration factors, $C_{UCB}$ is a constant, and $n(s_{1},s_{2},\cdots,s_{i})$
is the number of times the path $(s_{1},s_{2},\cdots,s_{i})$ has
been visited. 

Once the new categorical values have been sampled, we follow an \emph{update
strategy}, by updating $n(\bm{s})$ and $\bar{r}(\bm{s})$ where $\bm{s}$
denotes a vector like $(s_{1},s_{2},\cdots,s_{i})$, and the full
path $(s_{1},s_{2},\cdots,s_{L})$ corresponds to a categorical variable
value and propagates the updated value along the complete path
for each non-leaf node. The updated value is propagated along the
complete path for each non-leaf node. After a new
sample location is evaluated, the reward and visit count of all nodes along
the path of the tree that led to the sample are updated. For each node $s_{i}$ in the path, the visit count $n(s_{1},s_{2},\cdots,s_{i})$
is incremented by 1, and the average reward $\bar{r}(s_{1},s_{2},\cdots,s_{i})$
is updated to reflect the new sample's reward.

\subsection{Dynamic Model Selection}
\label{sec:kernel selection}
Based on the hybridization of tree and GP structures, we could claim
that our proposed model with the suggested search strategies can serve
as generalizations of most existing hybrid models listed in Table
\ref{tab:Model-unification-via}. Specifically, MCTS reduces to MAB
when $L=1$, and random sampling when the sequential sample count
$maxSampleSize$ in Algorithm \ref{alg:hybrid} is 0; the single mixed-variable
GP can be separated into independent continuous-variable GPs if block-diagonal
kernels that assume no correlation between categories are used for
$y$ (see Appendix \ref{subsec:Kernel-Design} for more detail). 
As a result, with some tailoring, our proposed model structure can reduce to MOSAIC, GP (skopt),
CoCaBO or EXP3BO.

Now, we introduce another important component in hybrid model, 
that is, the dynamic model selection. In the current paper, we consider the
commonly used kernels for GP surrogates, summarized in Table \ref{tab:Default-candidate-kernels}
and detailed in Appendix \ref{subsec:Kernel-Design}.

The transition from MCTS \st{and Bayesian strategies}  for parameter space
search to dynamic surrogate model selection can be understood as follows.
MCTS \st{and Bayesian} strategies determine sequential actions based on
average rewards and potential improvement either deterministically
or probabilistically. Similarly, in an online context like Bayesian
optimization, model selection involves choosing the best surrogate
model and the acquisition function, which balance the model's fit to observed
data (i.e., goodness-of-fit) and the potential improvement (i.e., acquisition).
Classical model selection criteria like AIC and BIC focus on the model's
fit but overlook the potential improvement. To address this, we propose
incorporating the acquisition function into model selection, as an analog to 
the \textquotedbl potential  improvement\textquotedbl{} concept in MCTS.
This leads to dynamic model selection, where we adaptively select
a covariance kernel from candidate kernels in each iteration, maximizing
optimization performance. The dynamic approach is partly evidenced by empirical studies like \citet{zeng2017progressive}.

\begin{table}
\centering{}%
\begin{tabular}{|c|c|c|}
\hline 
$k$  & $k_{\text{cat}}$  & $k_{\text{con}}$\tabularnewline
\hline 
\hline 
$k_{\text{cat}}+k_{\text{con}}$  & $k_{\text{MLP}}^{\sigma^{2},\sigma_{b}^{2},\sigma_{w}^{2}}$  & $k_{\text{Matern}}^{5/2,\ell}$\tabularnewline
\hline 
$k_{\text{cat}}+k_{\text{con}}$  & $k_{\text{Matern}}^{5/2,\ell}$  & $k_{\text{Matern}}^{5/2,\ell}$\tabularnewline
\hline 
$k_{\text{cat}}+k_{\text{con}}$  & $k_{\text{MLP}}^{\sigma^{2},\sigma_{b}^{2},\sigma_{w}^{2}}+k{}_{\text{Matern}}^{5/2,\ell}$  & $k_{\text{Matern}}^{5/2,\ell}$\tabularnewline
\hline 
$k_{\text{cat}}\times k_{\text{con}}$  & $k_{\text{MLP}}^{\sigma^{2},\sigma_{b}^{2},\sigma_{w}^{2}}$  & $k_{\text{Matern}}^{5/2,\ell}$\tabularnewline
\hline 
$k_{\text{cat}}+k_{\text{con}}+k_{\text{cat}}\times k_{\text{con}}$  & $k_{\text{MLP}}^{\sigma^{2},\sigma_{b}^{2},\sigma_{w}^{2}}$  & $k_{\text{Matern}}^{5/2,\ell}$\tabularnewline
\hline 
\end{tabular}\caption{\label{tab:Default-candidate-kernels} Default candidate kernels in
hybrid model, see Appendix \ref{subsec:Kernel-Design} for details.}
\end{table}

\subsubsection{Classical kernel selection criteria}

To maximize the performance of the model, we want to select a covariance
kernel $k$ from the candidates in Table \ref{tab:Default-candidate-kernels},
dynamically in each iteration during sequential sampling. 
Our framework allows to extend and add more candidate kernels at the cost of longer
computational time. This dynamic kernel selection in BO has not been
studied before and requires kernel selection criteria that incorporate
both how well a model fits the data and how much potential improvement
the model can lead to. We summarize a few classical kernel selection
criteria with a discussion about their drawbacks, followed by the proposed
rank-based criterion.

In an offline context, the model selection for GP regression falls
into two different categories: one category is based on the marginal
likelihood of the GP model; and the other category is the leave-one-out
cross-validation \citep{rasmussen_gaussian_2006}. In an online context,
the Bayesian information criterion (BIC) has been used in continuous
variable BO \citep{Malkomes2016Selection}: 
\begin{align}
\text{BIC}_{k} & =2\log\mathbb{P}_{k}(\bm{y})-n_{k}\log n\label{eq:BIC}
\end{align}
where $n_{k}$ is the number of kernel parameters of kernel $k$ and
$n$ is the number of samples in the sample response vector $\bm{y}$.
Other similar ones are
the Akaike information criterion (AIC), where $n_{k}\log n$ is replaced
by $2\cdot n_{k}$, and the Hannan--Quinn information criterion (HQC)
\citep{hannan1979determination}, where $n_{k}\log n$ is replaced
by $2n_{k}\log\left(\log n\right)$.

In addition, one can also use the (negative) log marginal likelihood
$\log\mathbb{P}_{k}(\bm{y})$ (indicating how well a model fits the
data) or the maximal acquisition function value $\mathbb{A}_{k}(\bm{y})\coloneqq\max_{x}\text{EI}(x,\mathbb{P}_{k}(\bm{y}))$
(indicating the potential improvement of the next sequential sample)
as the kernel selection criteria. However, the balance between log
likelihood, sample size, and acquisition function is critical for
these criteria to show stable and satisfactory results.

\subsubsection{\label{subsec:Kernel-Selection}Online kernel selection criteria}


\hlrev{
Kernel selection in GPs, traditionally a manual and expert-driven task, is computationally intensive} \citep{abdessalem2017automatic,bitzer2022structural}.
\hlrev{
Consider two extreme scenarios in kernel selection for Bayesian optimization -- one based purely on likelihood and the other on pure acquisition (optimism). The pure likelihood approach may lead to overly conservative model choices, possibly ignoring promising but less certain alternatives} (See Figure \ref{fig:Comparison-selection-critera}). \hlrev{Conversely, a pure acquisition-based approach risks favoring models that are overly optimistic but not necessarily accurate when away from the possibly local optimum, as in the case of a trivial surrogate that returns infinity everywhere in the design space in minimization contexts.

To choose among these candidate kernels in Table \ref{tab:Default-candidate-kernels}
for our surrogate model, a novel kernel selection criterion is proposed
next. In an online context for BO, we combine the strengths of optimism (highest acquisition) with likelihood measures for kernel selection. This combination is crucial to balance the explorative nature of acquisition with the data-based realism of likelihood assessments. 

Our decision to combine the ranks of these criteria, rather than their numerical values, is based on the need to mitigate the extremes of both approaches. Ranking allows us to integrate the optimistic explorative nature of acquisition with the grounded realism of likelihood assessments. This method ensures a more balanced and nuanced kernel selection process, and can be shown} (Figure \ref{fig:Comparison-selection-ker-Friedman8C}) \hlrev{to outperform fixed kernels in online  scenarios.
}

\textbf{Proposed rank-based kernel selection criterion.} We propose
a stable and effective selection criterion that well balances
likelihood and acquisition function by considering the rank $R_{\mathbb{P}}(k)$
of log likelihood $\log\mathbb{P}_{k}(\bm{y})$ and the rank $R_{\mathbb{A}}(k)$
of acquisition $\mathbb{A}_{k}(\bm{y})$ (larger quantities correspond
to higher ranks), both among all $k$. A generic formulation of such kernel selection criteria could take the form of $R_{\mathbb{P}}(k)+\alpha R_{\mathbb{A}}(k)$. In what follows, we focus on the following
\emph{rank-based kernel selection} criterion 
\begin{align}
R_{1/2}(k)=R_{\mathbb{P}}(k)+\frac{1}{2}R_{\mathbb{A}}(k) & ,\label{eq:C_k_custom}
\end{align}
where the factor $\frac{1}{2}$ puts less emphasis on the acquisition
function (analogous to BIC) since it is usually more sensitive than
log likelihood due to the fact that it incorporates uncertainty, and
also largely avoids tied ranks. Sometimes an adaptive coefficient
that puts more emphasis on the acquisition function as more samples
are generated can lead to more stable results, yielding e.g., an adaptive
selection criterion: 
\begin{align}
R_{ad}(k)=R_{\mathbb{P}}(k)+\frac{2\cdot i}{n}R_{\mathbb{A}}(k) & ,i=1,2,\cdots,n,\label{eq:C_k_custom-adaptive}
\end{align}
where $n$ is the deterministic sampling budget (i.e., $maxSampleSize$
in Algorithm \ref{alg:hybrid}) and $i$ means the $i$-th step among
all $n$ steps. In this paper, we use $R_{1/2}$ as the default criterion. 

The traditional kernel selection criteria (e.g., AIC, BIC) focus solely on the surrogate model's goodness-of-fit (like an ``exploitation'' part), making it apt for offline fixed datasets. Analogous to the Bayesian optimization's acquisition function, which balances exploration and exploitation, we suggest integrating an ``exploration'' component into the kernel selection criteria \eqref{eq:C_k_custom} and \eqref{eq:C_k_custom-adaptive}. Direct summation of these components can be problematic due to their differing magnitudes. Empirically, rank sums have proven more consistent. The coefficients $1/2$ and $2\cdot i/n$ serve as weights to balance exploration and exploitation during kernel selection. The former emphasizes the surrogate model's goodness-of-fit, while the latter, increasing with sequential sampling, accentuates exploration, especially when the goodness-of-fit near existing optima approaches perfection, as observed in standard Bayesian optimization.

We use the following example to illustrate how the rank-based criterion
$R_{1/2}$ works, where $1/2$ is chosen empirically, and refer to
Figure \ref{fig:Comparison-selection-critera} comparing different
selection criteria in the optimization context. The adaptive variant $R_{ad}$
with weights other than $1/2$ can be computed in a similar fashion. 
\begin{example*}
(Computation of $R_{1/2}$) Consider the candidate surrogate models
with kernels $k_{1},k_{2},k_{3}$, and suppose that we have already
computed their likelihoods based on the same set of samples $\bm{y}$.
Then, we can compute the relevant quantities below for kernel selection
purposes.

Based on the data below, we prefer the best-fitted model and the best
acquisition function, which is attained by a GP model with the kernel
$k_{1}$. The efficacy of the proposed kernel selection criterion
will be demonstrated with optimization examples in Section \ref{subsec:inactive_bench}.

\begin{adjustbox}{center}

\begin{tabular}{cccc}
kernel $k$  & $k_{1}$  & $k_{2}$  & $k_{3}$\tabularnewline
\midrule
\midrule 
$\log\mathbb{P}_{k}(\bm{y})$  & $\log\mathbb{P}_{k_{1}}(\bm{y})=2.6$  & $\log\mathbb{P}_{k_{2}}(\bm{y})=2.5$  & $\log\mathbb{P}_{k_{3}}(\bm{y})=-2.1$\tabularnewline
$R_{\mathbb{P}}(k)$  & 3  & 2  & 1\tabularnewline
\midrule 
$\mathbb{A}_{k}(\bm{y})$  & $\mathbb{A}_{k_{1}}(\bm{y})=2$  & $\mathbb{A}_{k_{2}}(\bm{y})=-1.5$  & $\mathbb{A}_{k_{3}}(\bm{y})=9.5$\tabularnewline
$R_{\mathbb{A}}(k)$  & 2  & 1  & 3\tabularnewline
\midrule 
$R_{1/2}$  & 4  & 2.5  & 2.5\tabularnewline
\end{tabular}

\end{adjustbox} 
\end{example*}
After generating a new categorical variable sample $\bm{s}$ via MCTS
(Section \ref{subsec:Monte Carlo-Tree-Search}), and selecting a mixed-variable
GP kernel $y(\bm{x})$ (Appendix \ref{subsec:Kernel-Design}) via
the rank-based selection criterion, one can search for new continuous
variable samples using the regular GP with the
chosen kernel, conditioned on the categorical sample $\bm{s}$. 

\subsection{Putting It Together: Algorithm for New Hybrid Model}\label{sec:put_together}
In our hybrid model\st{(both hybridM and hybridD)}, we consider the optimization for an objective function $f(\bm{x})=f(\bm{x}_{\text{cat}},\bm{x}_{\text{con}})$
with $\bm{x}_{\text{cat}}$ denoting the part of the categorical variables  and $\bm{x}_{\text{con}}$ denoting the part of the continuous vaiables,
for which the proposed hybridM 
model builds a mixed-variable GP but
separate the search phase using MCTS for $\bm{x}_{\text{cat}}$ and GP
for $\bm{x}_{\text{con}}$. Algorithm \ref{alg:hybrid} summarizes the complete pipeline when using the hybrid model, which efficiently constructs the categorical parameter
space as a tree structure, and associates each leaf node with a value
of $\bm{x}_{\text{cat}}$, i.e. a combination of categories.

Our hybrid models use tree and GP for the search phase, but only mixed-variable GP for the surrogate modeling. In MOSAIC \citep{rakotoarison_automated_2019} with fixed kernels, 
each categorical value combination is mapped to an independent GP. 
In our hybrid model, we employ a shared GP for all samples. During the search phase, categorical variables utilize a tree, while continuous ones use a GP. 
In the surrogate fitting phase, a single GP surrogate is applied, complemented by the novel online kernel selection. This unique construction leverages the efficiency of tree-search while ensuring adaptable GP model fitting with a list of mixed variable covariance kernels (See Appendix \ref{subsec:Kernel-Design}).

\begin{algorithm}[ht!]
\caption{\label{alg:hybrid}hybrid model algorithm with dynamic kernel selection and surrogate fitting.}
\KwData{$X_{n_{0},d}$ and $Y_{n_{0}}$ (data matrices consisting of $n_{0}$ pilot samples of $d$ coordinates in the variable, i.e., $d$ means the total number of continuous and categorical parts.)}
\KwIn{$maxSampleSize$ (the maximal number of evaluations of $f$), $L_{k}$ (the list of covariance kernels we want to consider).}
\KwResult{One tree structure with searching information, one GP surrogate model with trained covariance kernel.}

Initialize $X = X_{n_{0},d}$, $Y = Y_{n_{0}}$ and $i=n_{0}$ to be the sample size. Let $X_{\text{cat}}$ denote the categorical part of $X$\;
Train MCTS with $X_{\text{cat}}$ and $Y$\;
Train the GP surrogate with $X$ and $Y$, and a randomly selected kernel from the candidate kernels (see Section \ref{subsec:Kernel-Design})\;
\While{$i\leq maxSampleSize$}{
    Find the next categorical variable $\bm{x}_{\text{cat}}^{*}$ according to UCTS in Section \ref{subsec:Monte Carlo-Tree-Search} (i.e., hybridM) \st{or Bayesian strategy in Section 
    (i.e., hybridD)}\;
    \For{$k$ in $L_{k}$}{
        Train the GP surrogate $g_{k}$ with $X,Y$ and the covariance kernel $k$\ via likelihood maximization (MLE)\;
        Compute $R_{1/2}(k)$ defined in \eqref{eq:C_k_custom} (or \eqref{eq:C_k_custom-adaptive}) for the surrogate $g_{k}$ and kernel $k$\;
    }
    Find the kernel $k^{*}$ that maximizes \eqref{eq:C_k_custom} (or \eqref{eq:C_k_custom-adaptive})\;
    \tcc{The best kernel $k^{*}$ may be different for each iteration.}\;
    $\bm{x}_{\text{con}}^{*} = \arg\max_{\bm{x}_{\text{con}}}\text{EI}((\bm{x}_{\text{cat}}^{*},\bm{x}_{\text{con}}),g_{k^{*}})$\;
    $\bm{x}_{next}=(\bm{x}_{\text{cat}}^{*},\bm{x}_{\text{con}}^{*})$\;
    Append $\bm{x}_{next}$ to $X$, and $f(\bm{x}_{next})$ to $Y$\;
    Append $f(\bm{x}_{next})$ to the leaf node history and back-propagate\;
    Let $i=i+1$\;
}
\end{algorithm}

The algorithm assumes $n_{0}$ pilot samples, which are initial samples
that are observed prior to any sequential sampling or BO surrogate
training. We are allowed to search for $maxSampleSize$ sequential samples.
The hybrid model relies on three essential steps to generate the sequential
samples: 
\begin{enumerate}
\item Heuristic MCTS search for the categorical part in new samples and
update the MCTS via back-propagation (Lines 3 and 13 of Algorithm
\ref{alg:hybrid}). 
\item Dynamic kernel selection for the best mixed-variable GP surrogate
based on both the categorical and continuous parts of the next variable
(Lines 6-8 of Algorithm \ref{alg:hybrid}). 
\item Acquisition maximization of the continuous value conditioned on the
categorical values to add new samples to the GP surrogate. (Lines
9-13 of Algorithm \ref{alg:hybrid}). 
\end{enumerate}
We need to fit the model with all candidate kernels separately to
get different acquisition functions and selection criteria, which
is expensive. However, this step can be parallelized and we can ensure
that the selected kernel and surrogate will have the optimal balance
between exploring and exploiting.

\section{\label{sec:Experiments}Experiments and Analyses}

\subsection{\label{subsec:Synthetic-Functions}Synthetic Functions}

We compare the proposed hybrid  models \st{(both hybridM and hybridD)} against the following methods
we discussed above: TPE \citep{bergstra2011algorithms}, SMAC \citep{hutter_sequential_2011},
EXP3BO \citep{gopakumar_algorithmic_2018}, CoCaBO \citep{ru_bayesian_2020},
along with skopt optimizers \citep{headtim_scikitoptimize_2020} as
baselines. 
For skopt optimizers, skoptGP uses one-hot encoding for
categorical variables by default; skoptForest uses a plain random
forest estimator; skoptDummy randomly samples among all possible combinations
of categories. Recall that $C$ is the total number of possible combinations
of categorical values with $n_{c}$ categorical variables. In the
case of small $C$, we also provide roundrobin BO \citep{bergstra2011algorithms}
results, since roundrobin BO is clearly not favored
in the large $C$ case.

We compare the following methods for scientific applications (specific
versions in parentheses): TPE ($\mathtt{hyperopt==0.2.5}$), SMAC
($\mathtt{smac==1.2.0}$), skopt ($\mathtt{scikit-optimize==}$ $\mathtt{0.9.0}$
with $\mathtt{scikit-learn==1.0}$) 
and our hybrid models. Note that
we do not include the bandit-based methods (CoCaBO, EXP3BO, roundrobinMAB,
randomMAB) here as they require a predefined scaling factor in the
objective functions to avoid overflow problems in updating the reward.

The metric we use to express the computational budget (x-axis) is
the number of function evaluations (i.e., samples observed from the
black-box function). In the case of black-box functions, all function
evaluations are equally expensive in terms of time.  The performance
measure (y-axis) is the maximum objective black-box function value
(or minimum in minimization problems) that we observed so far at the sample
size. For a single batch, we can plot a curve for this ``best-so-far''
maximum value against the sample sizes explored. For multiple batches
(with different random seeds), we display the point-wise means and
standard deviations for multiple curves. In the line plot, the mean
of the optima is computed for each method; the standard deviation
is indicated by the vertical segments. We use 10 pilot samples and 90 sequential samples in all hybrid models and assign the same budget to other methods. \hlrev{Our study opted for a fixed number of pilot samples to maintain consistent comparisons across various mixed-variable methods. It is important to note that the scaling of pilot samples for higher dimensions across different methods remains an open problem in the field. Therefore, our approach reflects a compromise to balance methodological consistency with the diverse requirements of the different models in comparison following the existing comparison} \citep{ru_bayesian_2020}.

There are three kinds of important benchmarking functions that we
should be aware of in the mixed-variable setting. It is worth pointing
out that different approaches behave quite differently when applied
to different kinds of benchmark functions. This is caused by the different
correlations between categories, which is usually not explicitly modeled
in surrogates, especially for non-hybrid models.

\subsubsection{Categorical benchmarking}

{\small{}{}{}} 
\begin{figure}
\begin{tabular}{cc}
{\small{}{}\includegraphics[width=0.5\textwidth]{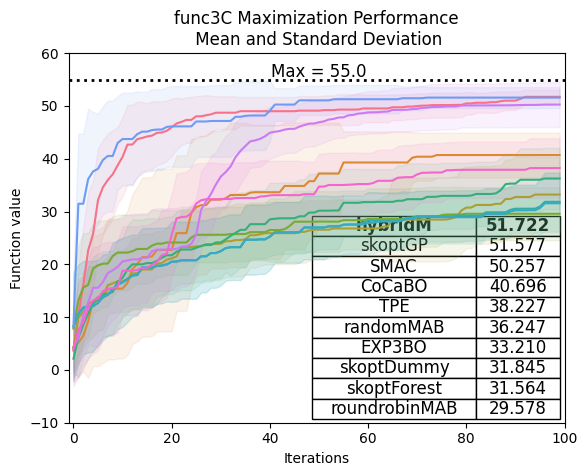}}  & {\small{}{}\includegraphics[width=0.5\textwidth]{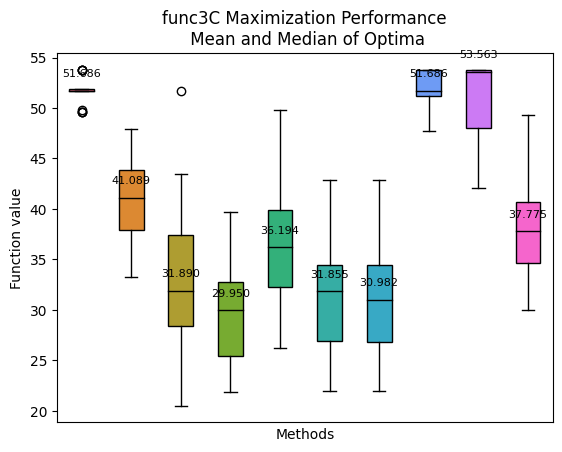}}\tabularnewline
{\small{}{}\includegraphics[width=0.5\textwidth]{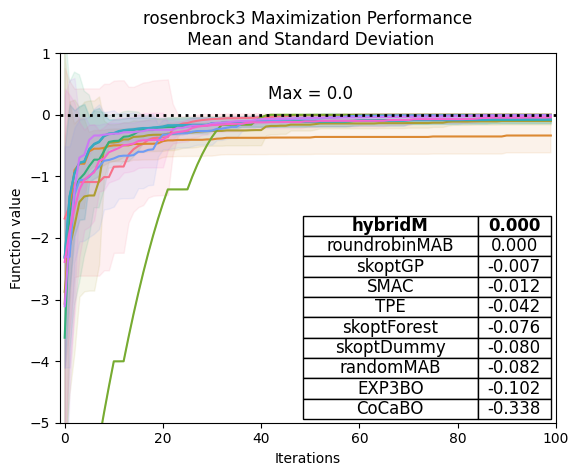}}  & {\small{}{}\includegraphics[width=0.5\textwidth]{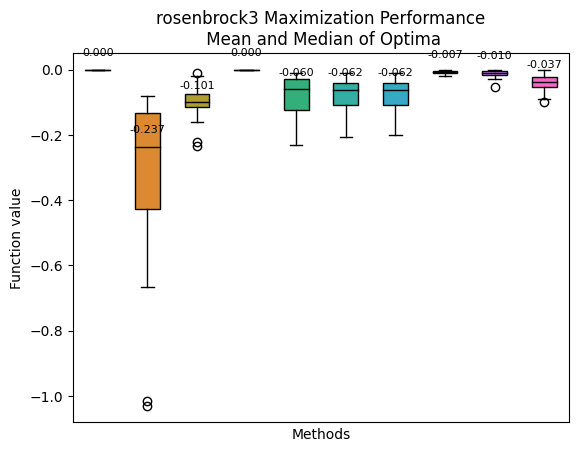}}\tabularnewline
{\small{}{}\includegraphics[width=0.5\textwidth]{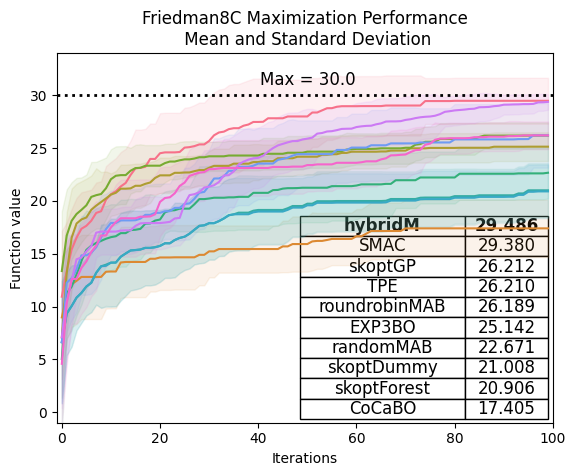}}  & {\small{}{}\includegraphics[width=0.5\textwidth]{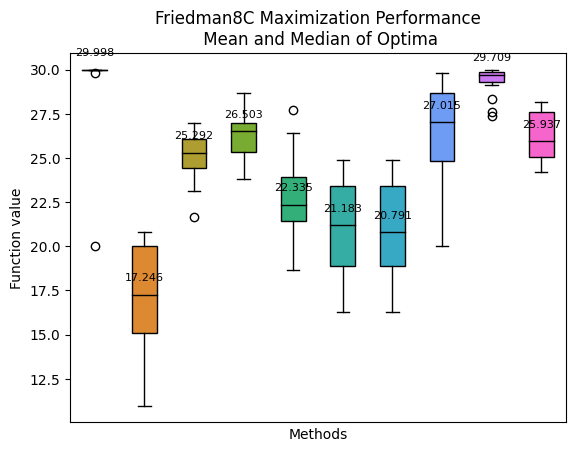}}\tabularnewline
\multicolumn{2}{c}{{\small{}{}\includegraphics[width=0.75\textwidth]{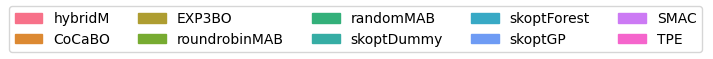}}}\tabularnewline
\end{tabular}

\vspace{-1.5em}
{\small{}{}\caption{{\small{}{}\label{fig:Comparison-joint-abc}} Comparison (with performance
line plot per budget and box plot of final optima) of performance
over 20 batches for: (row 1) a scaled version of the function func3C
in \citet{ru_bayesian_2020} with maximum 55; (row 2) a scaled version
of the discrete Rosenbrock function of 7 dimensions (4 continuous
variables in $[-5,5]$ and 3 categorical variables in $\{-5,-4,\cdots,4,5\}$)
in \citet{Malkomes2016Selection} with maximum 0; (row 3) function
\eqref{eq:Friedman-8C} with maximum 30.}
}{\small\par}

\end{figure}

The first kind of benchmark functions are created by assigning different
continuous-variable functions to different categories as shown in the first row of 
Figure \ref{fig:Comparison-joint-abc}. This kind of function
is represented by a (scaled version of) the function
func3C ($n_{c}=3,C=3\times5\times4=60$) from \citep{ru_bayesian_2020}.
The correlation between categories is not clear at all or non-existent.
The categorical benchmarking function is self-explanatory, like the
piece-wise constant functions involving discontinuity. This kind of
synthetic function mainly aims at checking if the mixed model is suitable
when the correlation is not necessarily continuous, which is the typical
setting the previous research focused on.

We can observe that hybrid models \st{(hybridM/hybridD)} are among the best models,
and enjoys a fast convergence rate. For the categorical benchmarking black-box
functions, hybridM needs a lower sampling budget to reach a place
near the optimal value and keeps improving until reaching the actual
optima. In the extreme case, this is similar to creating $C$ independent
surrogate models for each category (see Table \ref{tab:Model-unification-via}).
However, we only use one surrogate model for better usage under a
limited sampling budget, as explained in Appendix \ref{subsec:Kernel-Design}.

In this kind of benchmarking, the bandit strategy is quite suitable
for exploring the categories compared to covariance models. From our
empirical studies, hybridM   and the one-hot encoded GP (denoted by
skoptGP, where categorical is differentiated from continuous by encoding)
are all competitive. When $C$ is small, the one-hot encoding is
effective for the first kind of benchmarking.

\subsubsection{Integer-like benchmarking}

The second kind of benchmark function is constructed by discretizing
the functions with continuous variables only. That is, the continuous
variables are forced to take integer values only and treated as if
they are categorical. This kind of function is represented by discrete
Rosenbrock \citep{Malkomes2016Selection}, Ackley-cC \citep{nguyen_bayesian_2019}
and GP sample path functions considered by \citet{garrido-merchan_dealing_2020}.
The correlation between categories is induced by the original dependence
between continuous variables.

The well-known Rosenbrock function \citep{simulationlib} can be scaled
and defined for an arbitrary dimensional domain as: 
\begin{align}
f(\bm{x}) & =\frac{-1}{10000}\sum_{i=1}^{d-1}\left[100(\bm{x}_{i+1}-\bm{x}_{i}^{2})^{2}+(\bm{x}_{i}-1)^{2}\right],[-5,5]^{d}\subset\mathbb{R}^{d}\rightarrow\mathbb{R}.\label{eq:discrete rosenbrock_rosenbrock}
\end{align}
For example, when $d=7$ we may keep the first 4 dimensions continuous.
However, for the last 3 dimensions, instead of letting $\bm{x}_{5},\bm{x}_{6},\bm{x}_{7}$
vary continuously in $[-5,5]$, we force these variables to assume
values in the set $\{-5,-4,\cdots,4,5\}$ of 11 integers as categorical
values. These integers are taken as categorical encodings ($n_{c}=3,C=11\times11\times11=1331$).
This kind of discretization is another way of creating mixed-variable
benchmarking functions.

From the second row of Figure \ref{fig:Comparison-joint-abc}, we can observe that
our hybridM model has a slow start, but eventually reaches the actual
optima as bandit-based surrogate models like roundrobinMAB. 
The variances of all the surrogate models
considered are relatively low, which echoes the results in \citet{Malkomes2016Selection}.
In this kind of benchmarking, the bandit strategy is still reasonably
good in exploring the categories, but clearly less efficient compared
to covariance models. This is because the correlations between categories
are induced by discretizing the continuous domain, and before discretizing,
the relation is already well modeled by (stationary) kernels.

\subsubsection{Inactive benchmarking}

\label{subsec:inactive_bench}

The third kind of benchmark function contains one or more ``inactive''
categorical or continuous variables. In other words, some of the variables
do not affect the function value at all. \st{This sort of benchmarking
is ignored by previous studies on categorical variables, but arises
naturally in applications like screening, sensitivity analysis, and
computer experiments.} \hlrev{We highlight the challenges in identifying inactive variables in black-box optimization, which is a task not easily addressed by existing methods like post-hoc sensitivity analysis. Although our method does not directly address variable selection, emphasizing these difficulties underlines the critical need for advanced BO methods to discern and manage inactive variables.}

\st{This kind of synthetic function mainly aims at checking if the mixed
model can handle some inactive variables, and avoid the redundant
dimensions brought by these inactive variables.} A representative
example of this kind is the Friedman function. The Friedman function
is defined in such a way that not all categorical variables are active;
this function is a modified version of the well-known ``inactive
function'' by \citet{friedman_multivariate_1991} and has been extensively
studied in different contexts of variable selections \citep{chipman_bart_2010}.
We modify this function by appending several more inactive variables
to the Friedman-8C function: 
\begin{align}
f(\bm{x}) & =10\sin(\pi\bm{x}_{1}\bm{x}_{2})\cdot\bm{1}(\bm{x}_{7}=0)+20(\bm{x}_{3}-0.5)^{2}\label{eq:Friedman-8C}\\
 & +10\bm{x}_{4}\cdot\bm{1}(\bm{x}_{9}=0)-10\bm{x}_{4}\cdot\bm{1}(\bm{x}_{9}=1)+5\bm{x}_{4}\cdot\bm{1}(\bm{x}_{9}=2)+5\bm{x}_{5}\nonumber \\
f: & \bm{x}\in[0,1]^{6}\times\{0,1,2\}\times\{0,1,2,3,4\}\times\{0,1,2\}\times\{0,1,2,3\}^{3}\times\{0,1\}^{2}\rightarrow\mathbb{R}.\nonumber 
\end{align}
\hlrev{
The function comprises 6 continuous and 8 categorical variables, with limited combinations due to $\bm{x}_{10},\cdots,\bm{x}_{14}$ being restricted to ${0,1,2,3}$ ($n_{c}=8,C=11520$). Its non-stationarity, particularly as $\bm{x}_{9}=1$ shows different correlations with $\bm{x}_{9}=0$ and 2, makes it unsuitable for stationary kernels and well-suited for our non-stationary categorical kernel described in Appendix} \ref{subsec:Kernel-Design}.

\begin{figure}[ht!]
\includegraphics[width=1.\textwidth]{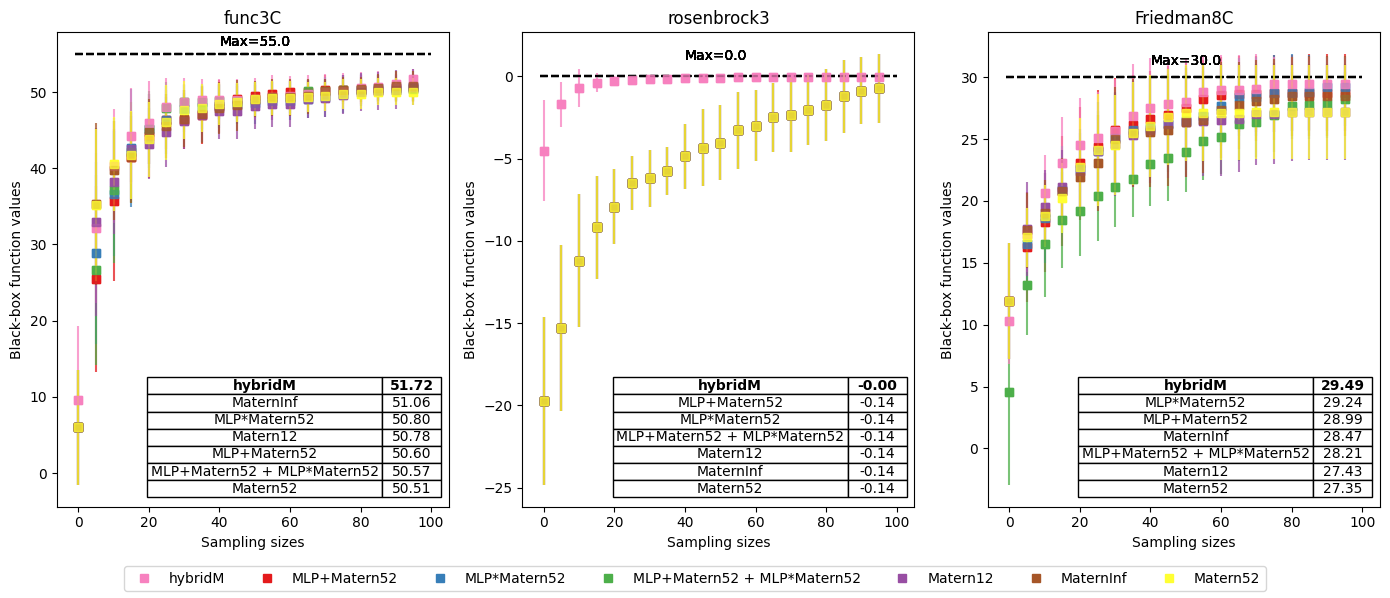}

\caption{\label{fig:Comparison-selection-ker-Friedman8C}Comparison of performance
between different but fixed kernels of GP surrogate with the selection
criterion \ref{eq:C_k_custom} and fixed kernel GP surrogates on the
functions func3C in \citet{ru_bayesian_2020}, \eqref{eq:discrete rosenbrock_rosenbrock}and \eqref{eq:Friedman-8C} over 20 batches. The actual maximum
is shown by dashed lines. 
}
\end{figure}

\begin{figure}[ht!]
\includegraphics[width=1.\textwidth]{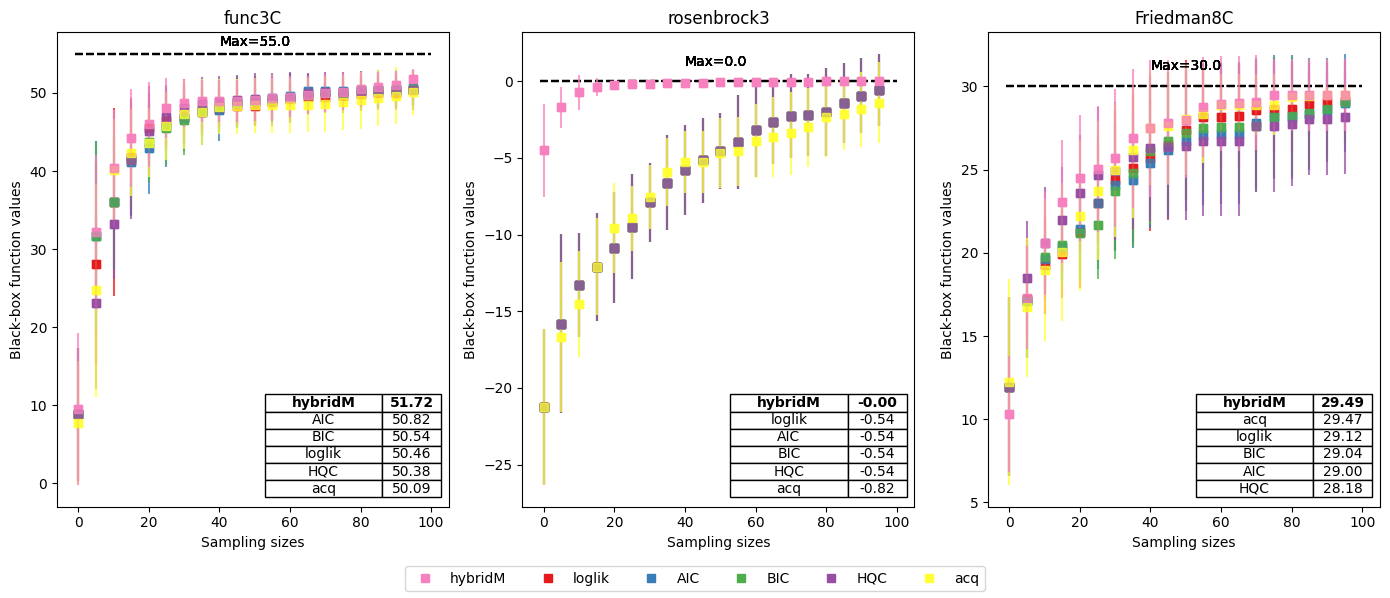}
\caption{\label{fig:Comparison-selection-critera}Comparison of performance
between the hybrid models with different selection criteria: acq (acquisition
function only), AIC, BIC, HQC, loglik (log likelihood) and $R_{1/2}$
in \eqref{eq:C_k_custom}, 
 on the functions func3C in \citet{ru_bayesian_2020}, \eqref{eq:discrete rosenbrock_rosenbrock}and \eqref{eq:Friedman-8C} over 20 batches. The actual maximum
is shown by dashed lines.}
\end{figure}

\hlrev{
Testing the Friedman8C function (Figure} \ref{fig:Comparison-joint-abc}), \hlrev{our hybridM model surpasses others in convergence and final optima,} \st{with hybridD} \hlrev{ followed by SMAC. The non-stationary MLP kernel amplifies the bandit's effectiveness, focusing on active categories. Most batches show hybridM reaching the true maximum efficiently. Stationary kernels often fail in mixed spaces, as our experiments confirm. HybridM excels with inactive variables or hierarchical spaces, and the bandit approach is a subset of hybrid models.

Using the Friedman-8C function} (\eqref{eq:Friedman-8C}), \hlrev{we illustrate the dynamic kernel selection's superiority in hybrid models} (Figure \ref{fig:Comparison-selection-ker-Friedman8C}). \hlrev{It enhances convergence and optima quality, with our expanded kernel families and a shared GP surrogate modeling both continuous and categorical variables. Figure} \ref{fig:Comparison-selection-critera} \hlrev{shows our rank-based criterion outperforming others in kernel selection} (Section \ref{subsec:Kernel-Selection}).

To sum up, the hybrid model 
has the best performance among all tested methods
on the benchmark functions of the first and third kinds in the sense
that it exhibits both fast or competitive convergence and best average
optima. For the second kind of benchmarking function, the hybrid model still
exhibits comparable average optima, but does not have the fastest
convergence rate (e.g., compared to skoptGP and skoptForest).

\subsection{\label{subsec:Scientific-Applications}Scientific Applications}

Next, we switch to expensive black-box functions arising in machine
learning and scientific computing applications. As we mentioned earlier,
the sampling budget is limited due to the fact that each sample is
costly to evaluate. The computational times for surrogate models are
neligible compared to the evaluation time of black-box functions.

We cannot standardize the reward, since the black-box function cannot
be bounded and scaled reasonably without knowing its range. \hlrev{Rescaling of reward functions have been known to be non-trivial and task-dependent} \citep{fouche2019scaling,lykouris2020bandits}. In addition, our hybrid model does not need standardization along the tree structure, in contrast to the bandits.

\subsubsection{Neural network regression hyper-parameter tuning }

\label{sec:Hyperparameter-Table} The feed-forward neural network
is used for a regression task. The regression task is to use the 13
observed variates in the Boston Housing dataset \citep{harrison1978hedonic}
to predict the response variable of the housing price (positive continuous
variable).

We take the negative mean square error (MSE) between the predictor and the observed response
as our objective function and set the task to tune the hyper-parameters
of the regression network among the range described in Table \ref{tab:Different-hyper-parameters-in}
to maximize the negative MSE. In our study, there is a collection
of multivariate regression data sets from the public repository with heterogeneity
and a red-shift dataset \citep{hrluo_2022e,elsken2019neural} from
cosmology simulations. The 3D-HST galaxy dataset \citep{skelton20143d}
underwent preprocessing: invalid redshifts and negative flux values
were removed; flux was converted to magnitude; and the data was split
into an 80-20 training-test ratio. The target for regression was the
z\_best column, indicating redshift.

\begin{table}
\begin{centering}
\begin{tabular}{|c|c|c|}
\hline 
Type  & Hyper-parameter  & Values\tabularnewline
\hline 
\hline 
\multirow{2}{*}{Categorical} & Number of Layers  & $\{3,4,5\}$\tablefootnote{the number of layers includes the input/output layer but not the dropout
layer.}\tabularnewline
\cline{2-3} \cline{3-3} 
 & Activation for Layers  & $\{\tan h,\text{ReLU},\text{sigmoid},\text{linear}\}$ \tablefootnote{all dense layers share the same activation function.}\tabularnewline
\hline 
\multirow{4}{*}{Continuous } & Layer Sizes  & $[4,16]\cap\mathbb{Z}$\tablefootnote{all dense layers share the same size.}\tabularnewline
\cline{2-3} \cline{3-3} 
 & Initial Learning Rate  & $10^{[-5,0]}$\tabularnewline
\cline{2-3} \cline{3-3} 
 & Layer Batch Size  & $[8,64]\cap\mathbb{Z}$\tabularnewline
\cline{2-3} \cline{3-3} 
 & Dropout Rate for Layers  & $[0.0,0.5]$\tabularnewline
\hline 
\end{tabular}
\par\end{centering}
\caption{\label{tab:Different-hyper-parameters-in}Different hyper-parameters
in the regression neural network.}
\end{table}

\begin{figure}
\includegraphics[width=1.\textwidth]{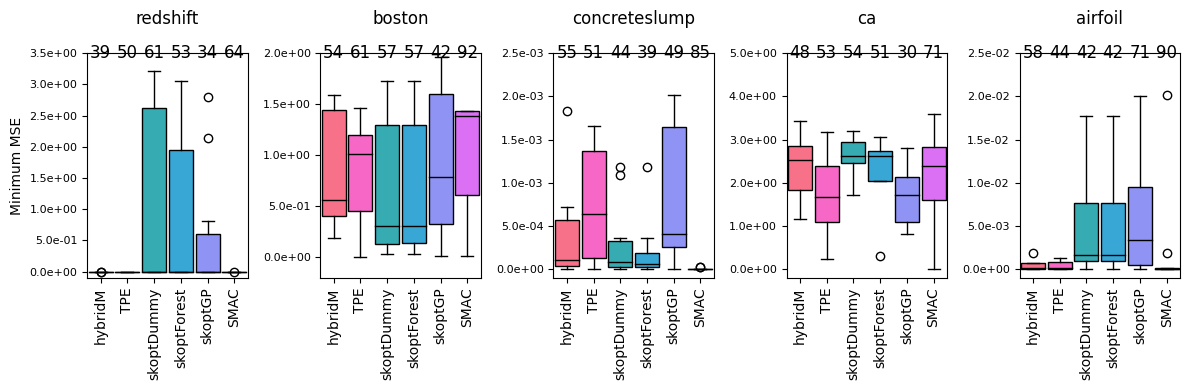} \caption{\label{fig:Comparison-NN} Comparison boxplots of performance between
different methods on the different tuning methods on the regression
neural network specified in Table \ref{tab:Different-hyper-parameters-in}
for datasets in Table \ref{tab:Data-format} over 10 repeated batches.
The theoretical minimum in-sample MSE is 0. We also display the average
iterations needed to attain optima at the top of each box. }
\end{figure}

\begin{table}[ht!]
\begin{centering}
%
\begin{tabular}{|c|c|c|c|}
\hline 
name & $n$ (sample size)  & $d$ (dimension) & source\tabularnewline
\hline 
\hline 
Galaxy redshift & 4173 & 30 & \citet{skelton20143d}\tabularnewline
\hline 
Boston housing & 506 & 13 & StatLib\tablefootnote{http://lib.stat.cmu.edu/datasets/boston}\tabularnewline
\hline 
Concrete slump test & 103 & 10 & uci\_ml\tablefootnote{https://archive.ics.uci.edu/dataset/182/concrete+slump+test}\tabularnewline
\hline 
California housing & 20640 & 8 & scikit-learn\tablefootnote{https://scikit-learn.org/stable/modules/generated/sklearn.datasets.fetch\_california\_housing.html}\tabularnewline
\hline 
Airfoil self-noise & 1503 & 6 & uci\_ml\tablefootnote{https://archive.ics.uci.edu/dataset/291/airfoil+self+noise}\tabularnewline
\hline 
\end{tabular}
\par\end{centering}
\caption{\label{tab:Data-format}Data format and source for the neural network
regression task.}
\end{table}

Figure \ref{fig:Comparison-NN} indicates that our methods have  attained  (near-)best optima on redshift, airfoil and concreteslump. Hybrid models also have 
reasonable convergence rates on the boston and california housing datasets, with comparable optima to the rest tuners. 
This echoes the observation of \citet{rakotoarison_automated_2019}
that a tree-based search is better able to handle larger categorical spaces.
For the galaxy redshift dataset, we found that the usual tuners from
skopt (not addressing the categorical nature of the hyper-parameters)
do not have good performance. TPE, SMAC and hybrid models show reasonable
tuning results, by taking categorical structure into consideration
(especially in airfoil and redshift). But SMAC usually takes more
budgets to attain the optimum, compared to relatively quick convergence
shown by TPE and hybrid models. When we have some apriori information,
the tree structure in TPE and hybrid models could help in searching  categorical space
efficiently. In this tuning neural network example, we do not claim
generic improvement but the robustness (compared to large variation
of skoptForest, skoptGP) and efficiency (compared to slow exploration
of SMAC, skoptDummy) of the hybrid models. It remains a nontrivial
problem how our tuning performance in this simple neural network problem can be generalized 
to generic neural architecture search \citep{elsken2019neural}.

After the initial comparison and examining the optimal configuration
found by the tuning methods, we are more informed that: 3-layer models
are not performing well compared to 4- and 5-layer models; 4-layer
models are not performing well when the activation function is sigmoid.
All models do not behave well for linear activation functions. It
is hard to incorporate these pieces of information into SMAC and TPE.
In the skopt framework, the typical way is to assign a very small
reward value (e.g., negative infinity) to these combination of parameters.
In hybrid models, the learned reward history \st{or Dirichlet posterior} 
allows us to transfer the information more easily compared to other
tuners. The hybridM and hrbridMD do not always find similar optima,
but often they will provide similar convergence rates. 

\subsubsection{STRUMPACK for three-dimensional Poisson problem}

\begin{figure}
\includegraphics[width=0.9\textwidth]{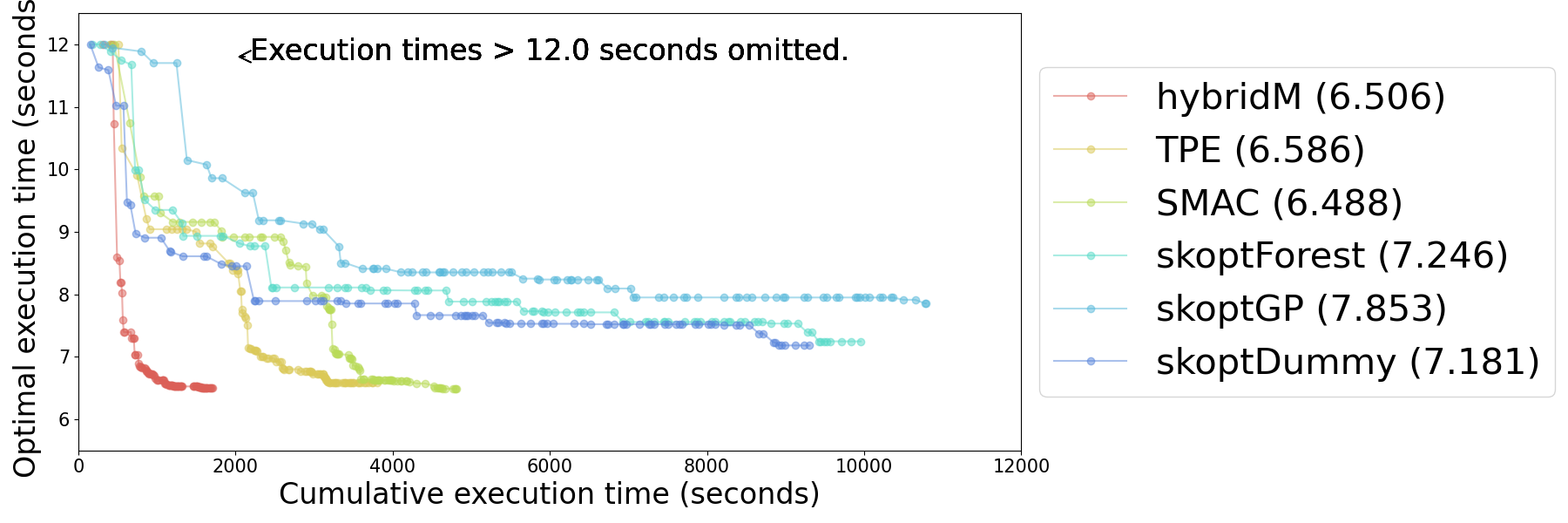}
\caption{\label{fig:Comparison-STRUMPACK-realtime} Comparison of performance
between different methods on the STRUMPACK for Poisson3d with grid
size 100. We display the actual (averaged over 10 batches) cumulative
execution time used for each sequential sample, and each line contains
100 sequential samples as in Figure \ref{fig:Comparison-STRUMPACK}. }
\end{figure}

In this computationally expensive experiment, we use the STRUMPACK
\citep{ghysels2016efficient,doecode_37094} sparse solver as a black-box
application with several categorical variables. STRUMPACK is a high-performance
numerical library for structured matrix factorization and we set the
tuning problem to be solving as fast as possible a 3-dimensional Poisson
equation on a $100\times100\times100$ regular grid, leading to a
sparse matrix of dimension $1\mathrm{M}\times1\mathrm{M}$. We consider
5 tuning-parameters with 3 categorical variables and 2 continuous
variables as shown in Table \ref{tab:Different-hyper-parameters-in-2}.
Each function evaluation requires 10 to 500 seconds using 8 Haswell
nodes of the Cori
, a Cray XC40 machine, at NERSC in Lawrence Berkeley National Laboratory.
For each tuner, we execute 10 batches with 100 sequential samples
due to a computational time limit (see Figure \ref{fig:Comparison-STRUMPACK-realtime}).
We measure the ``tuning budget'' by recording the actual cumulative
execution time in the function evaluation.

\begin{table}
\begin{adjustbox}{center} 
\begin{centering}
\begin{tabular}{|c|c|c|}
\hline 
Type  & Parameter  & Values\tabularnewline
\hline 
\hline 
\multirow{3}{*}{Categorical} & Fill-in Reduction Ordering & $\{'metis','parmetis','geometric'\}$\tabularnewline
\cline{2-3} \cline{3-3} 
 & Compression Algorithm  & $\{'hss','hodbf','blr'\}$\tabularnewline
\cline{2-3} \cline{3-3} 
 & Minimum Compressed Separator Size  & $\{2,3,4,5\}\times1000$\tabularnewline
\hline 
\multirow{2}{*}{Continuous} & Leaf Size in Compression  & $2^5,2^6,\cdots,2^9$\tabularnewline
\cline{2-3} \cline{3-3} 
 & Compression Tolerance  & $10^{-6},\cdots,10^{-1}$\tabularnewline
\hline 
\end{tabular}
\par\end{centering}
\end{adjustbox}

\caption{\label{tab:Different-hyper-parameters-in-2}Different parameters in
the STRUMPACK for Poisson3d with grid size 100.\\
*For the continuous variables, we treat their exponents as continuous and then round to the nearest integer.}
\end{table}

\begin{figure}[t]
\centering 
\includegraphics[width=0.65\textwidth]{figs/STRUMPACK_new2_accumulative_time_actual_time}
\includegraphics[width=0.65\textwidth]{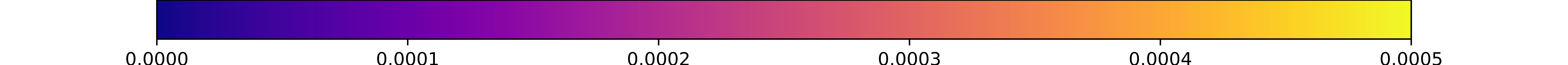}

\caption{\label{fig:Comparison-STRUMPACK-x-distribution} The histogram displaying
the density of the cumulative execution time (x-axis, seconds) against
actual execution time (y-axis, seconds) for each sample configuration
in the STRUMPACK application, visited by each method, over 10 batches.
We also display the actual cumulative time for each batch to finish
100 sequential samples in white (seconds).}
\end{figure}

The main time complexity falls on the execution of STRUMPACK, therefore,
we provide the plot of optimal execution time against the average
cumulative time taken by the STRUMPACK in Figure \ref{fig:Comparison-STRUMPACK-realtime}.
We can see that hybridM attains the optimal execution time quickly,
using only 50\% of the overall time to achieve the same or better
optima compared to the other models. And hybridM concentrates within
10\% of the optimal configuration at least 4 times faster than the
other model (Figures \ref{fig:Comparison-STRUMPACK-realtime} and
\ref{fig:Comparison-STRUMPACK-x-distribution}.) The other interesting
observation is that simple skopt models reach a reasonable optimal
execution time faster than TPE and SMAC. This can be further confirmed
by the histogram indicating the exploration pattern in Figure \ref{fig:Comparison-STRUMPACK-x-distribution}.
The better surrogate method should spend most of the sample budget
in exploring the configurations that take less execution time per
execution. In this figure, we saw that hybridM, TPE and SMAC spend
most of the sequential samples with near-optimal execution time, while
the simple skopt models explore sequential samples with worse execution
times. Some badly chosen parameter configurations can take the STRUMPACK
up to 500 seconds to complete, so the tuning cost is relatively high
even for 100 sequential samples but still far lower than an exhaustive
search.

\hlrev{
Although in our application the application execution takes the most time, the time complxeity of hybrid models is not worse than that of a regular GP surrogate used in BO.
The time complexity of the each selection at the node of a tree algorithm
can be computed as $O(m)$ since the UCB selection criterion takes
$O(1)$, where $m$ is the number of sequential children nodes. Since
the number $n_{c}$ of categories is the same as the number of layers
in the tree and the $i$-th category has $C_{i}$ categorical variables,
we need $O(n\cdot\sum_{i=1}^{n_{c}}C_{i})= O(n\cdot n_{c}\cdot\max_{i}C_{i})$
complexity for MCTS search for all $n$ sequential samples, in contrast to $O(n\cdot\prod_{i=1}^{n_{c}}C_{i})$ complexity for a MAB. At each leaf node, we need to fit $K$ different GP surrogate models
and select the best kernel based on our criterion. This step requires
time complexity is expressed by $O(Kn^{3})$, where we use all $n$
samples for the joint GP surrogate, different from MOSAIC. The total time complexity of our hybrid model is $O(n\cdot n_{c}\cdot\max_{i}C_{i}+Kn^{3})$,
bounded from above by a cubic term in $n$, which is the same as regular
GP.
}

This experiment reveals an interesting phenomenon: The simple models
(e.g., skoptGP and skoptForest) may attain a reasonable optimum relatively
quickly but fail to improve further. Advanced models (e.g., TPE and
SMAC) may be slow in achieving the same optimal level but keep improving
as samples accumulate. In stark contrast, hybridM is unique in the
sense that it would automatically select among simple and advanced
surrogate models with different kernels as the sampling happens, which
empirically balances between convergence rates and optimality.

\begin{figure}[h!]
\includegraphics[width=0.75\textwidth]{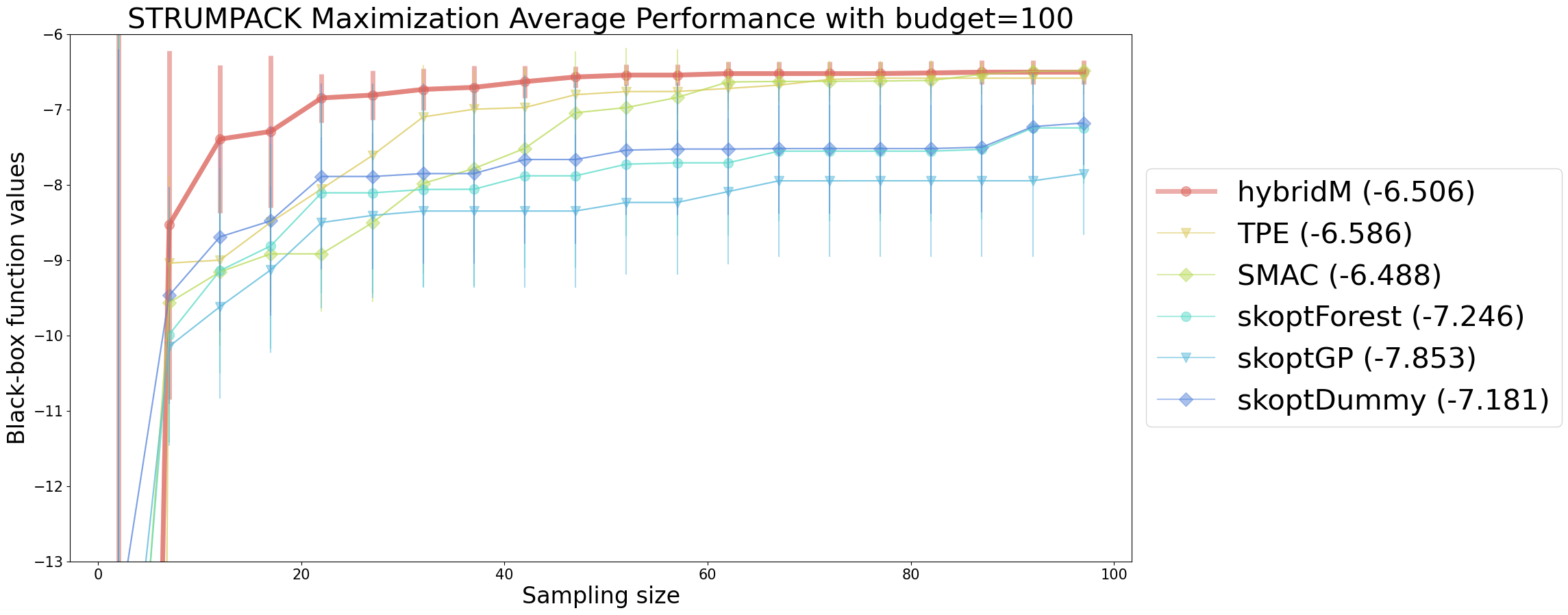}\\
 \includegraphics[width=0.75\textwidth]{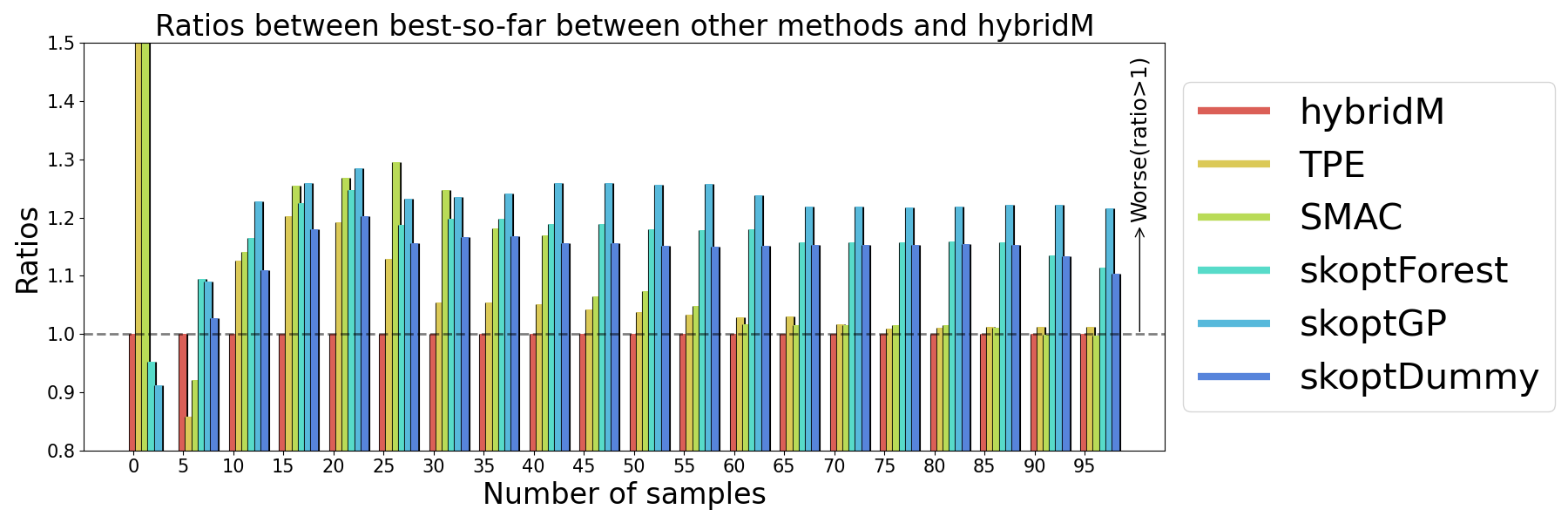}

\caption{\label{fig:Comparison-STRUMPACK}Comparison of performance between
different methods on the STRUMPACK for Poisson3d with grid size 100.
Note that we use negative optimal execution time in the upper panel
to match the line plot format in the previous applications.}
\end{figure}

\section{\label{sec:Conclusion}Contributions and Future Work}

This paper presents an efficient mixed-variable BO framework, using hybrid models (hybridM) that leverage novel MCTS \st{and Bayesian} strategies for searching
the categorical variable space and GP for searching the continuous
variable space. Hybridization of tree and GP is a unified BO framework that encompasses several
state-of-the-art mixed-variable optimization methods, leading to efficient search for categorical space. 
Our approach
distinguishes itself by incorporating an improved dynamic kernel selection
criteria, featuring a novel family of kernels well-suited for mixed
variables, and leveraging tree to reduce the categorical
search space. 

The key contribution of this chain of thought is the innovative integration
of concepts from MCTS \st{and Bayesian strategies}  into dynamic model selection
for Bayesian optimization \hlrev{using a dependent GP, where existing methods (e.g., MOSAIC) cannot handle limited sample per category}. This approach introduces a novel perspective
on model selection for online data, which traditionally focuses on the goodness-of-fit
of the model to observed data. By incorporating the acquisition function,
which represents the potential improvement, we extend the model selection
criteria to consider not only how well the model fits the data, but
also how much potential improvement the model can lead to. This dynamic
model selection process, which adaptively selects the best covariance
kernel from candidate kernels in each iteration, enhances Bayesian optimization performance.

We identified different types of benchmarking problem and provide
up-to-date benchmarks comparing these state-of-the-art BO methods
for mixed variables. Moreover, we applied the proposed algorithm to
ML and HPC applications to demonstrate its effectiveness compared
to most state-of-the-art mixed-variable BO methods. \hlrev{And these innovations are combined into a unified software pipeline.}

As mentioned in Appendix \ref{sec:Bayesian Update Strategies}, \hlrev{Bayesian strategies} will learn posterior
information through the updated reward functions or Dirichlet posteriors
at nodes. This opens the door to performing transfer learning across
tasks in mixed-variable tuning problems \citep{sid-lakhdar_multitask_2019,liao2023efficient}.
We recognize the limitations in \citet{tesauro2012bayesian} \hlrev{regarding the central limit theorem analog of the UCB reward, which is not fully Bayesian due to the inappropriate continuous Gaussian prior for the categorical space. Therefore, we aim to introduce a more suitable prior, potentially enhancing performance under a fully Bayesian framework.}
Future work for the proposed hybrid model for mixed-variable optimization
includes fully Bayesian MCTS strategies, automatic tree level operations,
and extensions to the multi-objective optimization context In addition,
we propose to develop tree operations that can handle generic constraints
for categorical variables as well.

\section*{Acknowledgements}

We gratefully acknowledge the Exascale Computing Project
(17-SC-20-SC), a collaborative effort of the U.S. Department of Energy
Office of Science and the National Nuclear Security Administration.
We used resources of the National Energy Research Scientific Computing
Center (NERSC), a U.S. Department of Energy Office of Science User
Facility operated under Contract No. DE-AC02-05CH11231. We stored
our code at \url{https://github.com/gptune/hybridMinimization} .

We sincerely thank Riley J. Murray and Rahul Jain for additional experiments
for randomized Kaczmarz algorithms and constructive suggestions for
our hybrid model on various applications.

We are grateful to the editor, the AE, and two anonymous reviewers for constructive
comments and suggestions that have significantly improved the article.
{\small{}{}{}  \bibliographystyle{chicagoa}
\bibliography{refs}
 }{\small\par}

\appendix
\newpage
\section*{Supplementary Materials}
\section{\label{sec:Bayesian Update Strategies}Bayesian Update Strategies}

In contrast to deterministic draws by maximizing \eqref{eq:UCB policy}
in MCTS, the proposed Bayesian update strategy involves sampling a
probability vector $\bm{p}$ from the Dirichlet distribution at each
node, and then using this probability~vector to define and draw from
a multinomial distribution. This multinomial draw selects the next
child node to visit and hence determines the categorical part along
the path. In the update, parameters of the Dirichlet distribution
are updated based on the rewards obtained from the search, allowing
the search to dynamically adapt its strategy based on the results
of previous searches.

The Dirichlet distribution is a multivariate generalization of the
Beta distribution \citep{tesauro2012bayesian} and is particularly
suitable for modeling categorical data or proportions. It is often
used as a prior distribution in Bayesian statistics, especially in
problems involving multinomial distributions due to their conjugacy.
The Dirichlet distribution is parameterized by a vector $\bm{\alpha}$
of positive reals, which can be interpreted as pseudocounts for the
categories of the Multinomial distribution. Given a vector of parameters
$\bm{\alpha}=(\alpha_{1},\alpha_{2},\ldots,\alpha_{K})$, a random
vector $\bm{p}=(p_{1},p_{2},\ldots,p_{K})$ follows a Dirichlet distribution
if its probability density function is given by:

\begin{equation}
f(\bm{p}\mid\bm{\alpha})=\frac{1}{B(\bm{\alpha})}\prod_{i=1}^{K}p_{i}^{\alpha_{i}-1}
\end{equation}
where $B(\bm{\alpha})$ is the multivariate Beta function. The vector
$\bm{p}$ represents the probabilities of choosing different $K$
actions and the reward vector $\bm{r}=\bm{r}(a)$ (deterministically depending on action $a$ as below) represents the $K$ different
rewards corresponding to different actions. The prior, likelihood and posterior can be expressed as follows: 
\begin{align}\mathbb{P}(\bm{p}\mid\bm{\alpha}) & \propto p_{1}^{\alpha_{1}-1}\cdots p_{K}{}^{\alpha_{K}-1}\sim\text{Dir}(\bm{\alpha}),\\
\mathbb{P}(a\mid\bm{p},\bm{\alpha}) & \sim\text{Multi}(\bm{p}),\label{eq:Dirichletupdate}\\
\mathbb{P}(\bm{p}\mid\bm{r}(a),\bm{\alpha}) & =\frac{\mathbb{P}(\bm{r}(a),\bm{p}\mid \bm{\alpha})}{\mathbb{P}(\bm{r}(a)\mid \bm{\alpha})}=\frac{\mathbb{P}(a\mid\bm{p},\bm{\alpha})\mathbb{P}(\bm{p}\mid\bm{\alpha})}{\mathbb{P}(\bm{r}(a)\mid \bm{\alpha})}\text{ via Bayes Theorem}\label{eq:Dirichlet update}\\
 & \propto\mathbb{P}(a\mid\bm{p},\bm{\alpha})\mathbb{P}(\bm{p}\mid\bm{\alpha}),\nonumber\\
 & \propto p_{1}^{r_{1}+\alpha_{1}-1}\cdots p_{K}{}^{r_{K}+\alpha_{K}-1}\propto\text{Dir}(\bm{\alpha}_{\bm{s}}+\bm{r}),\nonumber
\end{align}
where $\text{Dir}(\bm{\alpha})$ and $\text{Multi}(\bm{p})$
are Dirichlet and multinomial distributions with respective parameters. 

In the context of the tree search strategy, the Dirichlet distribution
is used to model the uncertainty about the probabilities of selecting
each child node. The parameters $\bm{\alpha}$ at each node associated
with the Dirichlet distribution are updated each time a node is visited,
based on the rewards obtained from the search. This allows the search
to dynamically adapt its strategy based on the results of previous
searches, favoring paths that have led to higher rewards in the past.
We initialize the Dirichlet prior as a non-informative prior following
the interpretation of \citet{sethuraman1982convergence} that the
base measure $\bm{\alpha}(\cdot)$ is the prior observed sample size
and forcing $\bm{\alpha}$ to tend to zero means no prior information.

The novel Bayesian update strategy is as follows: We sample a $K$-vector
$\bm{p}$ from the Dirichlet distribution $\text{Dir}(\bm{\alpha}_{\bm{s}})$
at the parent node along the path $\bm{s}=(s_1,s_2,\cdots,s_i)$ where the node $s_i$ has $K$ possible categorical values as children, then a $K$-vector $a$
from the multinomial distribution $\text{Multi}(\bm{p})$. Then we
obtain a reward $r(a)$ from the playout vector $a$ and update the
reward distribution. 

Next, we convert the reward $r(a)$ into a reward vector $\bm{r}=(r_{1},r_{2},\cdots,r_{K})$
and update the Dirichlet distribution into $\text{Dir}(\bm{\alpha}_{\bm{s}}+\bm{r})$.
One primitive strategy that considers only exploitation is to set the
reward vector $\bm{r}$ to binary vector with all entries equal
to zero, except for the entry corresponding to the action $a$ that
has the largest average reward, which is set to a constant 1 (0-1
reward vector). 
\begin{equation}
\bm{r}\coloneqq(r_{1},r_{2},\cdots,r_{K})=\left(\delta_{1},\delta_{2},\cdots,\delta_{K}\right),\label{eq:0-1 reward}
\end{equation}
where the notation $\delta_{k}$ is 1 if the average reward at children
$k$ is the largest among all of its siblings, otherwise 0. A strictly
positive variant of this reward $\left(1+\delta_{1},1+\delta_{2},\cdots,1+\delta_{K}\right)$
can be used. However, this leads to empirically slow convergence,
and we choose to include the denominator $N_{k}$ to balance out the
exploration-exploitation trade-off as shown in \eqref{eq:UCB policy}.
\\
Our reward vector is defined to be 
\begin{equation}
\bm{r}\coloneqq(r_{1},r_{2},\cdots,r_{K})=\left(\frac{1+\delta_{1}}{N_{1}},\frac{1+\delta_{2}}{N_{2}},\cdots,\frac{1+\delta_{K}}{N_{K}}\right),\label{eq:our reward}
\end{equation}
where the denominator $N_{k}=n(s_{1},s_{2},\cdots,s_{i})$
is the number of times the tree path $(s_{1},s_{2},\cdots,s_{i})$
has been visited following the previous notations. 

We have to point out that although the 0-1 reward vector seems to
be a natural choice, our alternative reward vector \eqref{eq:our reward}
converges faster in almost all experiments. Besides the 0-1 reward,
another natural reward function is to take a proportional distribution:
the reward $r(a)$ is distributed among the entries of the reward
vector $\bm{r}$ in proportion to some predefined weights. These weights
could represent the prior belief about the importance or relevance
of each action. This strategy allows for a more nuanced assignment
of rewards to actions, which can be useful in complex environments
where the contribution of each action to the reward is not binary,
but varies in degree. Our alternative \eqref{eq:our reward} can be
considered as a modification of 0-1 reward, and we show its advantages via experiments but leave theoretical justification as future work.

In the view of reward function, this strategy effectively assigns
the entire reward to the action that was taken, and no reward to the
other actions. This strategy ensures that the search dynamically adapts
its strategy based on the results of previous searches, favoring paths
that have led to higher rewards in the past. 

The UCTS strategy (denoted as hybridM) in Section \ref{subsec:Monte Carlo-Tree-Search}
uses the formula \eqref{eq:UCB policy} to guide the search in a deterministic
manner. The UCB formula balances exploration (visiting fewer explored
nodes) and exploitation (visiting nodes that have high average reward).
The UCB value of a node is calculated based on the average reward
of the node and a confidence interval that depends on the number of
times the node and its parent have been visited. The node with the
highest UCB value \eqref{eq:UCB policy} is selected for the next visit. It uses a more deterministic
approach (although we introduce $\epsilon$-greedy) that balances
exploration and exploitation based on a deterministic maximization.

The Bayesian Dirichlet-Multinomial strategy (denoted as hybridD) in Section \ref{sec:Bayesian Update Strategies}
uses a Dirichlet-Multinomial conjugacy \eqref{eq:Dirichlet update} to generate a sample of probabilities
for each child node. This sample is then used to perform a multinomial
draw to select the next node to visit. The parameters of the Dirichlet
distribution are updated each time a node is visited, based on the
rewards \eqref{eq:our reward} obtained from the search. It also takes uncertainty into account
but may converge slowly in situations where the reward structure is
complex and the optimal path is not clear from the start.

\section{\label{subsec:Kernel-Design}Candidate Kernels for the Mixed-variable
GP }
In what follows, we use a synthetic
function to explain these steps: 
\begin{align}
f(\bm{x}_{1},\bm{x}_{2},\bm{x}_{3})=\pi\cdot\bm{1}(\bm{x}_{1}=0)+e\cdot\bm{1}(\bm{x}_{1}=1)+0.618\cdot\bm{1}(\bm{x}_{2}=2)+\bm{x}_{3}\label{eq:simple example}\\
y(\bm{x}_{1},\bm{x}_{2},\bm{x}_{3})=f(\bm{x}_{1},\bm{x}_{2},\bm{x}_{3})+\epsilon
\end{align}
Here $\bm{x}_{1}\in\{0,1\}$, $\bm{x}_{2}\in\{2,3\}$, and $\bm{x}_{3}\in\mathbb{R}$,
hence $\bm{x}_{\text{cat}}=(\bm{x}_{1},\bm{x}_{2})$ and $\bm{x}_{\text{con}}=\bm{x}_{3}$.
$y$ is the desired mixed-variable GP surrogate for the objective
function $f$ with noise $\epsilon$ and the variable consisting of
$d=3$ coordinates in the notation of Algorithm \ref{alg:hybrid}.
$X=X_{n_{0},d}=\{\bm{x}_{1},\cdots,\bm{x}_{n_{0}}\}$ will consist
of the input variables from $n_{0}$ initial samples, where $Y=Y_{n_{0}}=\{f(\bm{x}_{1})+\epsilon,\cdots,f(\bm{x}_{n_{0}})+\epsilon\}$
are corresponding noisy observations drawn from the objective function,
following our assumptions.

{\small{}{}Gaussian process (GP) modeling is mainly used for building
surrogate models $g$ with continuous variables in Bayesian optimization
\citep{gramacy_surrogates_2020}. We briefly describe the procedure
in BO and refer our readers to \citet{shahriari_taking_2016} and
\citet{gramacy_surrogates_2020} for formal details. In the procedure
of optimization, we sequentially draw (expensive) samples from the
black-box function $f$ and update the surrogate model $g$ with the
samples drawn. For two sample locations, $\bm{u},\bm{u}'$, their
correlation is defined by a covariance kernel function $k(\bm{u},\bm{u}')\in\mathbb{R}$
as we illustrate below. Supposing that the the surrogate GP model
$g$ approximates the black-box function $f$ sufficiently well, we
can find the true optimum $f_{\max}=\max_{\bm{x}}f(\bm{x})$ of the
black-box function. 
{\small\par}

{\small{}{}The sequential samples are selected based on the fitted
surrogate model $g$; more precisely, we maximize the acquisition
functions based on the surrogate model $g$ to select the next sequential
sample. In this paper, we only consider the expected improvement (EI)
acquisition function \citep{shahriari_taking_2016}.}{\small\par}

{\small{}{}When applied to mixed-variable functions, the quality
of the GP surrogate depends on the covariance kernel that models both
categorical and continuous parts simultaneously. However, existing
kernels are developed mainly for continuous variables  \citep{ru_bayesian_2020}
and cannot be modified for categorical variables straightforwardly.
Among the variety of different kernels for both continuous and categorical
variables, several different observations have been made in the literature
arguing whether regular continuous kernels or specialized kernels
are suitable for categorical variables \citep{garrido-merchan_dealing_2020,karlsson2020continuous}.}{\small\par}

{\small{}{}The form of the single mixed-variable GP surrogate $y$
for the black-box function $f$ is performance critical and highly
problem dependent. For example, stationary kernels, both specialized
(e.g. CoCaBO) \citep{ru_bayesian_2020} and the usual ones (e.g.,
Matern 5/2, without encoding) \citep{karlsson2020continuous} have
been demonstrated to show advantages in the context of some mixed-variable
tuning, but worse in other scenarios. Besides the stationary kernels,
we extend our candidate kernels under consideration to include both
non-stationary kernels, which have been rarely used in the context
of mixed-variable BO, and composition kernels as explained below using
the example function \eqref{eq:simple example}.}{\small\par}

{\small{}{}Stationary kernels. This kind of kernel does not distinguish 
between continuous and encoded categorical variables, but is used
mainly for continuous BO. Kernels in this family only depend on distances
between variables, e.g. 
\begin{align}
k_{\text{Matern}}^{\nu,\ell}(\bm{u},\bm{u}')=\frac{1}{\Gamma(\nu)2^{\nu-1}}\Bigg(\frac{\sqrt{2\nu}}{\ell}\|\bm{u}-\bm{u}'\|\Bigg)^{\nu}B_{\nu}\Bigg(\frac{\sqrt{2\nu}}{\ell}\|\bm{u}-\bm{u}'\|\Bigg)\label{eq:matern}
\end{align}
where $B_{\nu}$ is the modified Bessel function of order $\nu$,
$\Gamma$ is the gamma function and $l$ is a kernel parameter \citep{rasmussen_gaussian_2006}.
In the function \eqref{eq:simple example}, $\bm{u}=(\bm{x}_{1},\bm{x}_{2})$
if the kernel is used for categorical variables, and $\bm{u}=\bm{x}_{3}$
if the kernel is used for continuous variables. Note that for a general
categorical variable, e.g., $\bm{u}=\{a,b,c\}$, we can use the default
encoding $\bm{u}=\{0,1,2\}$ in \eqref{eq:matern}. In this paper,
we use $\nu=5/2$, but different smoothness may affect the performance
of BO \citep{luo_nonsmooth_2021}.}{\small\par}

{\small{}{}Non-stationary kernels. For encoded categorical variables,
we include the following non-stationary MLP (multi-layer perceptron)
arc-sine kernel (a.k.a., arc-sine in (3.5) of \citet{hermans2012recurrent})
\begin{align}
k_{\text{MLP}}^{\sigma^{2},\sigma_{b}^{2},\sigma_{w}^{2}}(\bm{u},\bm{u}')=\sigma^{2}\frac{2}{\pi}\text{asin}\left(\frac{\sigma_{w}^{2}\bm{u}^{T}\bm{u}'+\sigma_{b}^{2}}{\sqrt{\sigma_{w}^{2}\bm{u}^{T}\bm{u}+\sigma_{b}^{2}+1}\sqrt{\sigma_{w}^{2}\bm{u}'^{T}\bm{u}'+\sigma_{b}^{2}+1}}\right),\label{eq:MLP kernel}
\end{align}
Here $\sigma^{2}$, $\sigma_{b}^{2}$, $\sigma_{w}^{2}$ are kernel
parameters, and $\bm{u}=(\bm{x}_{1},\bm{x}_{2})$ for the function
\eqref{eq:simple example}. We argue that this is more suitable than
the stationary overlapping kernel depending on the Hamming distance
between the categorical variables $\bm{u},\bm{u}'$ (CoCaBO \citep{ru_bayesian_2020}).
The usual stationary kernels relies on the definition of metrics,
however, it is hard to characterize even (linear) similarity between
categories using regular distance metrics \citep{cerda2018similarity}.
We hope that a non-stationary kernel along with suitable encoding
could alleviate this problem and improve the quality of the model,
and illustrate this via experiments.}{\small\par}

{\small{}{}As far as we know, non-stationary kernels are rarely studied
for the mixed-variable in BO context, and we want to include such
a kernel in our hybrid models. We choose the arc-sine kernel since
it introduces inner products between encodings, and describes conditional
embeddings and similarity between categories. The form of arc-sine
kernel is also closely related to the arc-distance kernel proposed
for conditional space BO \citep{swersky2014raiders}, and the arc-cosine
kernel proposed for similarity learning \citep{cho2009kernel}. We
select the arc-sine kernel for its implementation simplicity.}{\small\par}

{\small{}{}Composition kernels. We use the Matern 5/2 and MLP kernels
to develop a few composition kernels for mixed-variables using candidate
kernels, as shown in Table \ref{tab:Default-candidate-kernels}.}{\small\par}
{\small{}{}Summation kernel models the additive effect between the
continuous and categorical variables: 
\begin{equation}
k_{\text{sum}}(\bm{x},\bm{x}')=k_{\text{cat}}\left((\bm{x}_{1},\bm{x}_{2}),(\bm{x}'_{1},\bm{x}'_{2})\right)+k_{\text{con}}\left(\bm{x}_{3},\bm{x}'_{3}\right).\label{eq:sum kernel}
\end{equation}
Product kernel models the interaction effect between continuous and
categorical variables: 
\begin{equation}
k_{\text{product}}(\bm{x},\bm{x}')=k_{\text{cat}}\left((\bm{x}_{1},\bm{x}_{2}),(\bm{x}'_{1},\bm{x}'_{2})\right)\cdot k_{\text{con}}\left(\bm{x}_{3},\bm{x}'_{3}\right).\label{eq:product kernel}
\end{equation}
A mixture of summation and product kernels \citep{ru_bayesian_2020}
is commonly believed to take into account both additive and interaction
effects: 
\begin{align}
k_{\text{mix}}(\bm{x},\bm{x}')= & (1-\lambda)k_{\text{sum}}(\bm{x},\bm{x}')+\lambda k_{\text{product}}(\bm{x},\bm{x}'),\lambda\in[0,1].\label{eq:mix kernel}
\end{align}
}{\small\par}
but we do not use this family of kernels. 
\end{document}